\documentclass[final,5p,times,twocolumn]{elsarticle}
\usepackage[colorlinks,urlcolor=blue,linkcolor=blue,citecolor=blue]{hyperref}

\usepackage{booktabs}       
\usepackage{amsfonts}       
\usepackage{nicefrac}       
\usepackage{microtype}      
\usepackage{xcolor}         
\usepackage{algorithm, algorithmicx, algpseudocode}
\usepackage{amsmath}
\usepackage{amssymb}
\usepackage{amsthm}
\usepackage{graphics,graphicx}
\usepackage{bbding}
\usepackage{multirow}
\usepackage{subfigure}
\usepackage{float}

\newcommand{\R}{\mathbb{R}}
\newcommand{\Cp}{\mathcal{C}_{\phi}}

\newcommand{\bb}{\textbf{b}}
\newcommand{\bW}{\textbf{W}}

\newcommand{\bu}{\textbf{u}}

\newcommand{\T}{{\intercal}}

\newtheorem{lemma}{Lemma}

\newtheorem{proposition}{Proposition}
\newtheorem{definition}{Definition}

\journal{xxx}

\begin{document}

\begin{frontmatter}



\title{A Lightweight and Gradient-Stable Neural Layer}

\author[1,3]{Yueyao Yu} 
\author[2,3]{Yin Zhang}

\address[1]{School of Science and Engineering, The Chinese University
		of Hong Kong-Shenzhen, China}
\address[2]{School of Data Science, The Chinese University
		of Hong Kong-Shenzhen, China}
\address[3]{Shenzhen Research Institute of Big Data, China}
 
\begin{abstract}

To enhance resource efficiency and model deployability of neural networks, we propose a neural-layer architecture based on Householder weighting and absolute-value activating, called Householder-absolute neural layer or simply Han-layer.  Compared to a fully connected layer with $d$-neurons and $d$ outputs, a Han-layer reduces the number of parameters and the corresponding computational complexity from $O(d^2)$ to $O(d)$.  {The Han-layer structure guarantees that the Jacobian of the layer function is always orthogonal, thus ensuring gradient stability (i.e., free of gradient vanishing or exploding issues) for any Han-layer sub-networks.}   Extensive numerical experiments show that one can strategically use Han-layers to replace fully connected (FC) layers, reducing the number of model parameters while maintaining or even improving the generalization performance.  We will also showcase the capabilities of the Han-layer architecture on a few small stylized models, and discuss its current limitations.
\end{abstract}




\end{frontmatter}


\section{Introduction}

The advancement of neural networks has sparked a revolution across multiple disciplines. However, as models grow larger and larger, so do their demands for resources such as energy and computing costs.  For example, training a large language model~(LLM), with the number of parameters reaching hundreds of billions, can cost millions of 
U.S.~dollars~\cite{energyconsider}.  On the other hand,  most devices in real applications are resource-constrained, such as mobile or vehicle-mounted devices, making it hard to deploy over-sized models.  Consequently, there is a growing demand for leaner models with fewer parameters that can operate efficiently with adequate functionality~\cite{energyconsider, vitsurvey}. 

Although researchers have investigated various methods, such as model compressing and pruning, to reduce model sizes (see recent surveys~\cite{vitsurvey, liang2021pruning} and references therein), so far there is still a lack of a low-complexity (in term of parameter and operation counts) and gradient-stable layer architecture that can effectively replace fully connected layers under suitable conditions.  In this paper, we propose a new neural network layer called the Householder Absolute-value Neural layer (or Han-layer), which replaces the dense weight matrix in a fully connected layer with a Householder matrix and uses the absolute-value function as the activation function. 

Our choice of Householder matrix is motivated by two key factors. First, it is an orthogonal matrix, ensuring gradient stability during training.  Second, a $d \times d$ Householder matrix involves just $d$ parameters, thereby making it possible to yield a substantial reduction in the overall parameter count even when multiple Han-layers are used to replace a single fully connected layer.  On the other hand, our selection of  {absolute-value function, or ABS}, as the activation function is guided primarily by the consideration that the Jacobian matrix of ABS, applied element-wise, is a diagonal and orthogonal matrix (with $\pm 1$ on its diagonal).  Additionally, ABS is a piecewise linear function, like the popular ReLU function, with a low computational complexity.

 {
It is crucial to underscore the complementary roles played by Householder weighting and ABS activating.  The combination of the two ensures orthogonality of the Jacobian matrices for each and all layer functions (see Equation \eqref{G-matrix} below), regardless of the number of layers.  This orthogonality ensures gradient stability without needing to deploy conventional techniques such as normalizations and residual connections~\cite{batchnorm,weightnorm, resnet, residualnormpre}.  In addition, this very combination inherits the benefits of being lightweight (from Householder weighting) and being computationally efficient (from ABS activating).}



\subsection{Contributions}

We propose a new, lightweight and gradient-stable layer architecture called Han-layer, and carefully examine its properties.  We conduct extensive experiments to evaluate the capabilities of neural networks equipped with multiple Han-layers (or simply HanNets for brevity). In addition, since Han-layers are 1-Lipschitz continuous, they are naturally resistant to adversarial attacks to a degree. 

\begin{figure}[htp]
	\centering
	\subfigure[FCNet: 85.2\% test accuracy.]{
		\includegraphics[width=.11\textwidth,trim=110 40 100 60, clip]{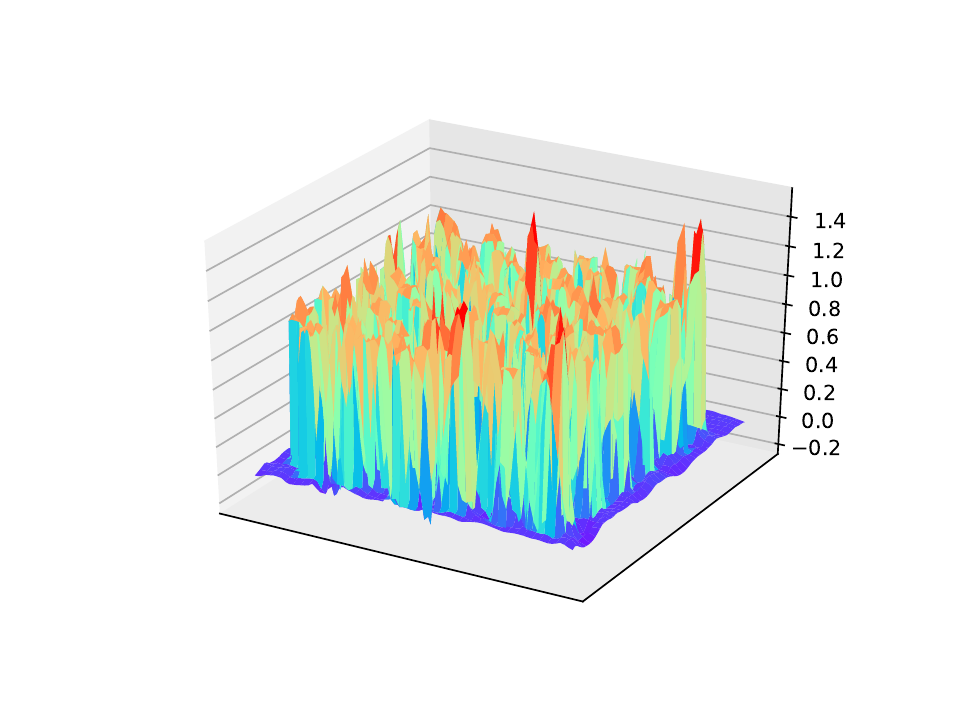}
		\includegraphics[width=.11\textwidth,trim=110 50 100 60, clip]{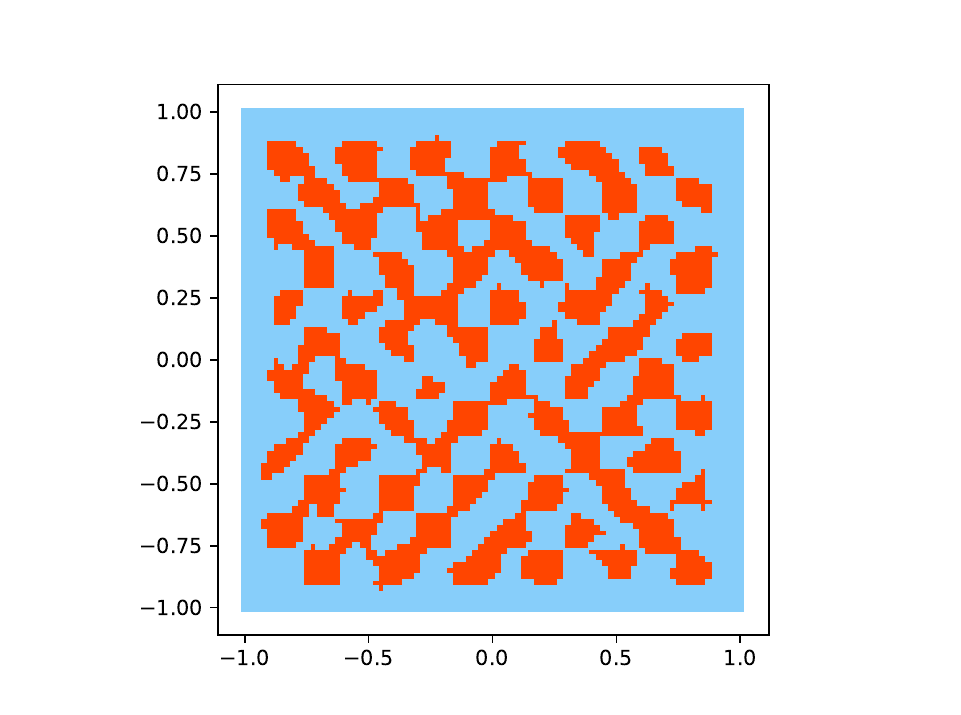}}
	\subfigure[HanNet: 99.5\% test accuracy.]{
		\includegraphics[width=.11\textwidth,trim=110 40 100 60, clip]{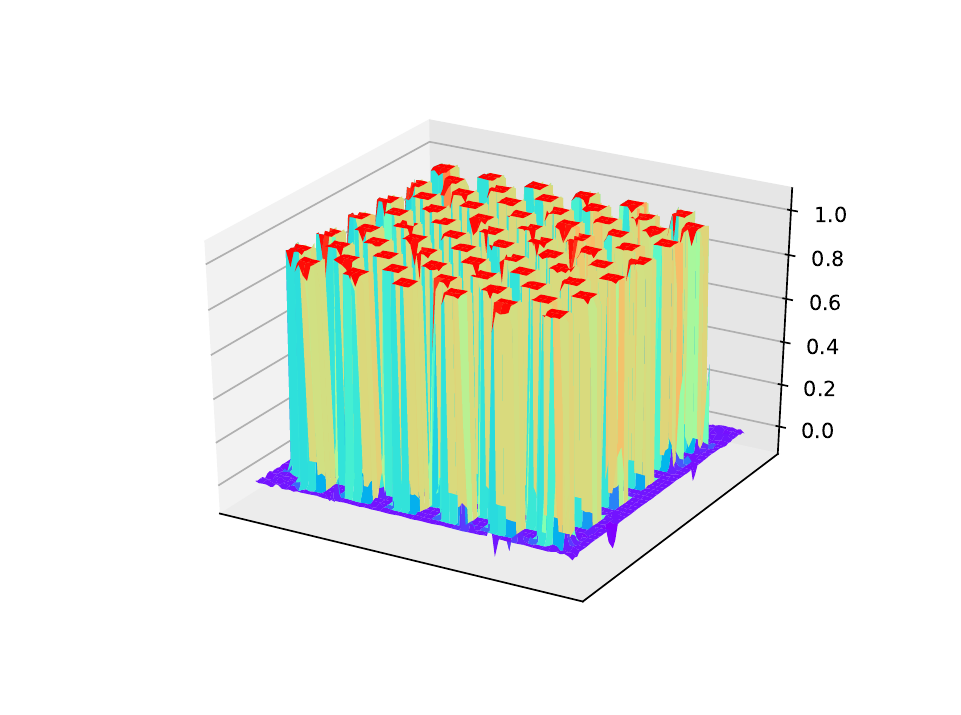}
		\includegraphics[width=.11\textwidth,trim=110 50 100 60, clip]{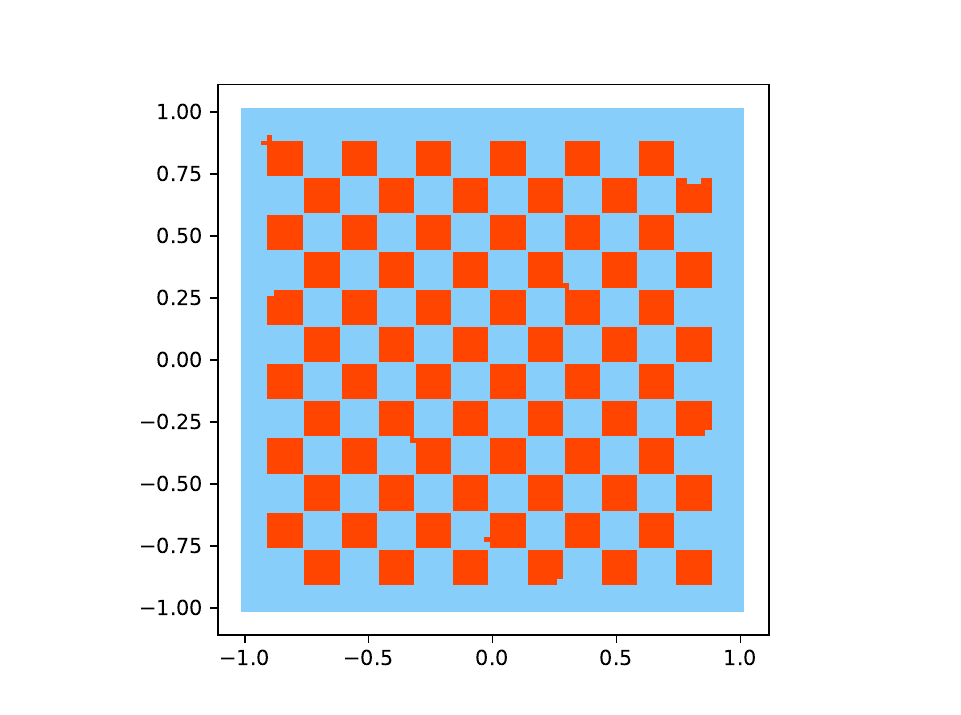}}
	\vspace{-.2cm}
	\caption{Landscapes and top views for FC and Han models on the checkerboard dataset: (a) FCNet, and (b) HanNet. }
	\label{fig:checkerboard}
\end{figure}

\begin{itemize}
	
	\item On several standard datasets for regression and image classification, our experiments indicate that HanNets can significantly reduce the number of model parameters, {even for models already characterized as lightweight,}  while maintaining, and in some cases improving, generalization performance. 
	
	\item On stylized small problems (see Section~\ref{result:Checkerboard}), our experiments show the existence of some structured data on which HanNets can greatly outperform conventional {Multi-layer Perceptrons~(MLPs)} in terms of generalization, as is demonstrated in Figure~\ref{fig:checkerboard} where the HanNet result is nearly perfect.

\end{itemize}

Even though our empirical results in this paper\footnote{For reproducibility, all the relevant codes are posted in GitHub at the link: \href{https://github.com/yyy32/HanNet}{https://github.com/yyy32/HanNet}.} are obtained from small- to moderate-scale models, they should be sufficient to demonstrate that the new Han-layer architecture represents a useful technique to be included in the existing toolbox of deep learning.  It may be particularly useful in building lightweight models, for example, in mobile applications.

\subsection{Notations}

We define a function $F_L$ from $\R^m$ to $\R^n$ realized by a deep neural network with depth $L$ as follows
\begin{equation}\label{def:FL}
	F_L(x,\bW,\bb) := (\psi_L\circ\cdots\circ \psi_1)(x),
\end{equation}
where
$\bW = \{W_1,... , W_L\}$ and $\bb = \{b_1, ... , b_L\}$ are the collections of weight matrices and bias vectors, respectively; $\psi_i$ is the $i$-th layer function 
with an activation function $\phi$, defined as
\begin{equation}\label{def:Gi}
	\psi_i(\cdot) \equiv \psi_i(\cdot,W_i,b_i) := \phi(W_i(\cdot) + b_i),
\end{equation}
where the scalar function $\phi(\cdot)$ is applied element-wise. The deep neural network (DNN) function $F_L(x,\bW,\bb)$ can be computed via forward propagation,
\begin{equation}\label{def:recursion}
	s_0 = x; \;\; z_k = W_ks_{k-1}+b_k, s_k = \phi(z_k), \;\; k = 1,\cdots,L. 
\end{equation}

For convenience of subsequent analysis, here we calculate the following Jacobian matrix involved in the gradient computation through back-propagation, denoted as the $G_L$-matrix,
\begin{equation}\label{G-matrix}
	G_L(x,\bW,\bb) \equiv
	\frac{\partial F_L(x,\bW,\bb)}{\partial x}
	= \prod_{k=L}^1 W_k \nabla\phi(z_k),
\end{equation}
where $W_k\nabla\phi(z_k)$ is the Jacobian matrix of the $k$-th layer, and $\nabla\phi(z_k)$ is the diagonal, Jacobian matrix of $\phi(\cdot)$ evaluated component-wise at a vector $z_k$.  The spectral norm of the $G_L$-matrix can serve as an indication whether gradient vanishing or exploding will happen.  Asymptotically speaking, as $L \rightarrow \infty$, gradient vanishing happens if $\|G_L\|_2 \rightarrow 0$ and gradient exploding happens when $\|G_L\|_2 \rightarrow \infty$. 

%
{We denote the Householder (reflection) matrix associated with a nonzero vector $u \in \mathbb{R}^d$ as
\begin{equation} \label{eq:H}
	H(u) = I - 2\frac{uu^{\T}}{u^{\T}u}, 
\end{equation}
where $I$ is the identity matrix of dimension $d$. $H(u)$ is both symmetric and orthogonal.  The number of parameters in $H(u)$ is $d$, an order of magnitude lower than that of a general $d$ by $d$ matrix.  }

\subsection{Related Work}

{The proposed Han-layer architecture combines ABS activating with Householder weighting based on  two central considerations: (1) guaranteed orthogonality of every layer Jacobian $H(u_k)\nabla\phi(z_k)$, and (2) low computational complexity. Consequently, the total Jacobian matrix of multiple Han-layers remains orthogonal since products of orthogonal matrices are orthogonal, ensuring gradient stability (i.e., no gradient exploding or vanishing) in any HanNet portion of a network.}

{Since the proposed combination of the two key components has not yet been previously investigated, in the following two subsections we will survey related work for ABS activating and Householder weighting separately.  We note that neither of the two techniques has been widely used, though neither is really new.}

\subsubsection{Absolute-Value Function for Activation}

 {
ABS stands out as one of the simplest {\it nonlinear} functions whose Jacobian matrices, whenever existing, are always orthogonal (see Sec.~\ref{sec:ABS}). In addition, it is piecewise linear and easy to apply at a low cost. In the literature, we found only a couple of other piecewise linear activation functions with orthogonal Jacobian matrices (when existing) which are named MaxMin and GroupSort and proposed in~\cite{groupsort}. Later on, we will compare ABS with these two in an ablation experiment, see Table~\ref{tab: ablation}.}
 
In earlier periods of neural network research, piecewise linear function ABS was used~\cite{batruni1991multilayer, lin1992canonical}. However, ABS failed to gain mainstream popularity and was seldom utilized.  For example it is not mentioned in this recent survey~\cite{jagtap2023important} of activation functions. Instead, ReLU has become the most widely used (piecewise linear) activation function, as demonstrated in~\cite{nair2010rectified}. The author in~\cite{karnewar2018aann} suggests using ABS activation in a ``bidirectional neuron" architecture based on interpretability considerations. More recently, in~\cite{zhang2021variability}, the authors show that under the same conditions, ABS can better resist the occurrence of  ``collapse to constant" and maintain network variability better than ReLU. Additionally, in~\cite{beknazaryan2021analytic}, the author demonstrates that neural networks with absolute value activation functions can $\epsilon$-approximate functions that are analytic on certain regions.

{Beside ABS and ReLU, there are many other piecewise linear activation functions studied in the literature, see the survey \cite{tao2022piecewise} and a specific example in \cite{xu2020efficient}.  Most previous studies on piecewise linear activation functions are not directly related to ABS activating since Jacobian orthogonality is out of their scope of considerations.}

\subsubsection{Householder Weight Matrices}

{In recurrent neural networks (RNN), products of Householder reflection matrices are utilized to represent transition matrices in order to alleviate issues related to gradient instability, including \cite{householderrnnefficient,zhang2018stabilizing} (see \cite{mathiasen2020one} for a related variant).   There are two major differences between the previous use of Householder weighting and ours: (1) we couple each Householder weight matrix with the ABS activation to ensure orthogonality of the layer Jacobian; (2) we only make use of a single Householder matrix per layer to create light-weight layers.
}

{Beside mitigating gradient instability, the sustained orthogonality (as opposed to at initialization only) of Householder matrices has been used to enhance learning quality in a number of application scenarios, such as in generative imaging~\cite{Song_2023_ICCV}, variational auto-encoders~\cite{tomczak2016improving}, model robustness~\cite{householderactivation} and speaker anonymization~\cite{miao2023speaker}.
}

{In addition to using Householder reflection matrices, other methods are proposed to enforce sustained orthogonality, exemplified by \cite{arjovsky2016unitary, vorontsov2017orthogonality, Wisdom2016full} where Householder, permutation, Cayley Transformation and/or other forms of orthogonal or unitary matrices are utilized (or combined) to form weight matrices.  Other works, such as \cite{wang2020orthogonal, jia2019orthogonal, brock2017neural}, try to keep weight matrices approximately orthogonal by various means, including using singular-value decomposition or regularization terms such as  $\|W^{\T} W-I\|_F^2$.  In \cite{jia2019orthogonal}, the authors demonstrate that orthogonal neural networks, leveraging local isometry properties, manifest superior generalization performance.
}

{Finally, orthogonal weight-matrix initializations have been the subject of theoretical and empirical investigations, such as in \cite{Hu2020Provable, huang2018orthogonal, pennington2018emergence, xiao2018dynamical}.  Even though initialization techniques do not maintain sustained weight orthogonality during training, they have proven to be useful in deep learning.
}


\section{Introduction to Han-layer}
\begin{figure*}[ht]
	\centering
		\includegraphics[width=.9\textwidth,trim=80 390 80 100, clip]{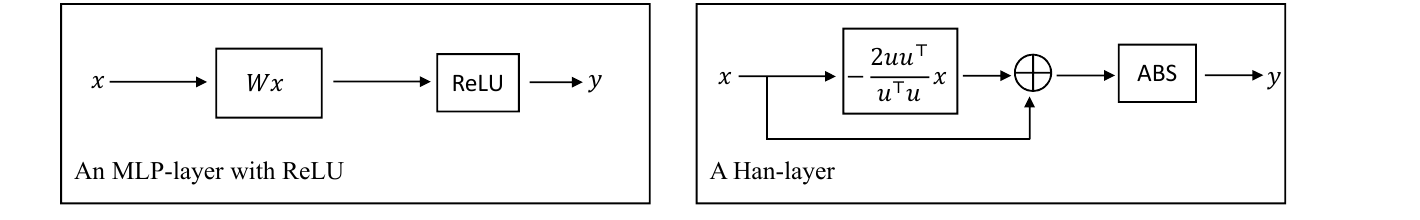}
	\vspace{-.2cm}
	\caption{Visualization of an MLP-layer with ReLU activation and a Han-layer with ABS activation. On the left, $W$ is $d \times d$ and $Wx$ requires $O(d^2)$ operations, while on the right $u$ is a nonzero $d$-vector and the multiplication with $x$ requires only $O(d)$ operations.  }
	\label{fig:visualization-mlp/han}
\end{figure*}
A Householder-absolute neural layer, or Han-layer, is composed of a Householder matrix followed by ABS activation, 
\begin{equation}
\varphi(x;u,b) := \mathbf{ABS} \left(H(u)x+b\right),
\end{equation}
which has a linear complexity. 
We will call a neural network mainly composed of Han-layers as a HanNet.   Additionally, Figure~\ref{fig:visualization-mlp/han} visually illustrates the distinction between a conventional MLP-layer and a Han-layer. It is interesting to note that our Han-layer can be seen as having an intrinsic residual connection.


\subsection{Absolute-value Activating}
\label{sec:ABS}
We motivate element-wise absolute-value activating by considering a class of functions defined below.
\begin{definition}
	A function $\phi(t): \R\rightarrow\R$ has Property~A if 
	\begin{enumerate}
		\item[A1.] $\phi$ is nonlinear and continuous in $\R$;
		\item[A2.] $\phi$ is differentiable in $\R$ except in a countable set $\Cp\subset\R$;
		\item[A3.] $\phi$ satisfies the following unit-derivative condition:
		\begin{equation}\label{unit-d}
			|\phi'(t)| = 1, ~~ \forall t \in \R\setminus\Cp.
		\end{equation}
	\end{enumerate}
\end{definition}
We restrict non-differentiability to a countable set to avoid unnecessary complications introduced by uncountable sets.

Nonlinearity, continuity and differentiability (almost everywhere) are universally required for activation functions with obvious or well-explained rationales~\cite{groupsort}.  Hence, conditions A1-A2 are natural.

It is easy to see that condition~A3 is the necessary and sufficient condition for the element-wise function $\phi(x): \R^d\rightarrow\R^d$ to have an orthogonal Jacobian matrix.  Therefore, to guarantee the orthogonality of the $G_L$-matrix in \eqref{G-matrix}, we propose to impose the unit-derivative condition~\eqref{unit-d} on element-wise activation functions.

\begin{lemma}
	Let $\phi: \R\rightarrow\R$ have Property~A and $|\Cp|$ denote the 
	cardinality of $\Cp$. Then $|\Cp| \ge 1$.  
\end{lemma}
We notice that $|\Cp|$ is the number of non-differentiable points.
This result follows immediately from the observation that if $\Cp$ is empty, then $\phi$ would be differentiable on the entire line with a constant derivative and thus must be linear which would fail condition~A1 (otherwise there would exist $t_0\in\R$ so that $\lim_{t\to t_0^{+}} \phi'(t)\neq \lim_{t\to t_0^{-}} \phi'(t)$).

\begin{lemma} \label{lemma:abs}
	Let $\phi: \R\rightarrow\R$ have Property~A.  Then $\phi$ is piecewise linear 
	so that each piece has the functional form $\phi(t) = \pm t + \mathrm{const}$ with a different constant .  
	In the case of the minimum cardinality $|\Cp|=1$ with 
	$\Cp=\{\alpha\}$, $\phi$ has the functional form
	\begin{equation}\label{abs2}
		\phi(t) = \pm|t - \alpha| + \mathrm{const}.
	\end{equation}
\end{lemma}
\begin{proof}
The first part is obvious.  When $\Cp=\{\alpha\}$, the functional form in each piece is $\phi = (t-\alpha)+\mathrm{const}$ or $-(t-\alpha)+\mathrm{const}$, then $\phi$ has the functional form $\phi(t) = \pm|t - \alpha| + \mathrm{const}$.
\end{proof}

\begin{figure*}[ht]
	\centering
	\subfigure[Approximate FC-1. Left to right: the original FC-1, the trained FC-1, the trained Han-3.]{
		\includegraphics[width=.25\textwidth,trim=80 30 70 30, clip]{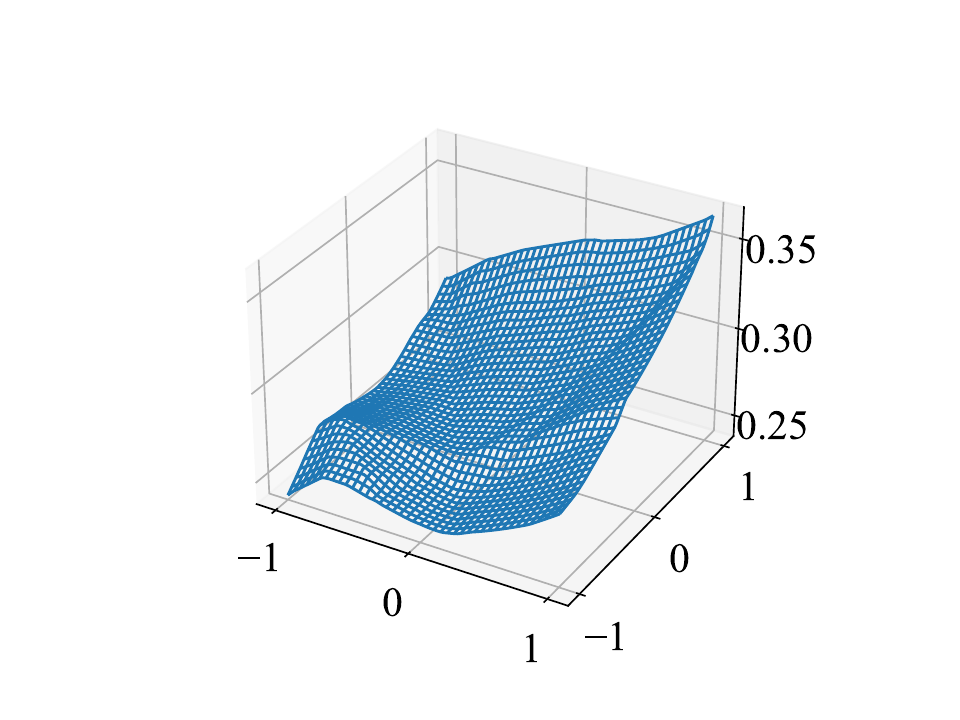}
		\includegraphics[width=.25\textwidth,trim=80 30 70 30, clip]{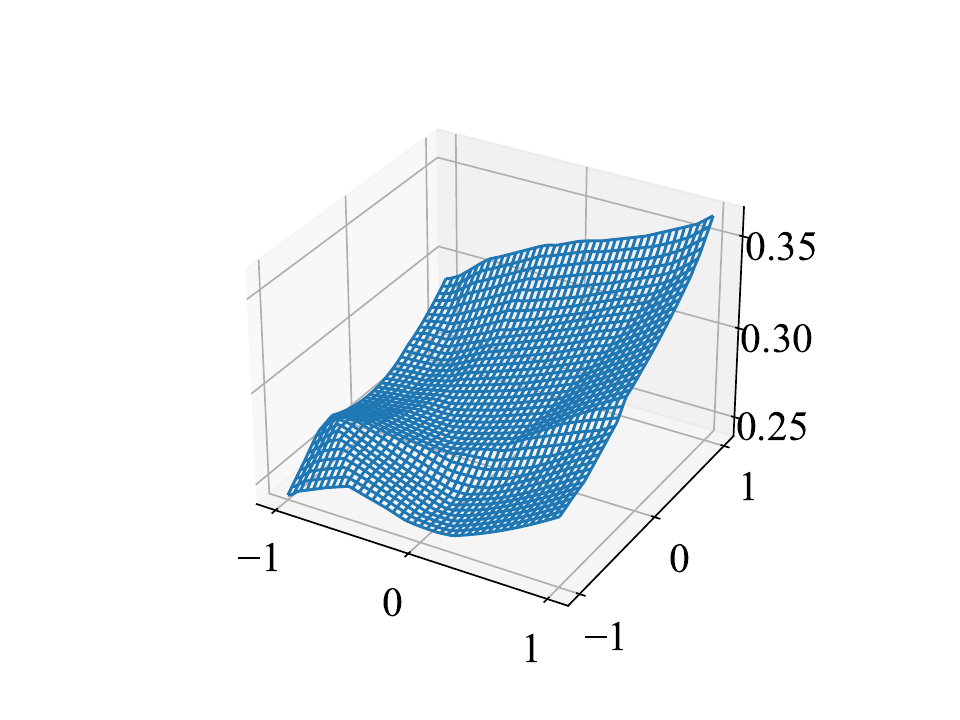}
		\includegraphics[width=.25\textwidth,trim=80 30 70 30, clip]{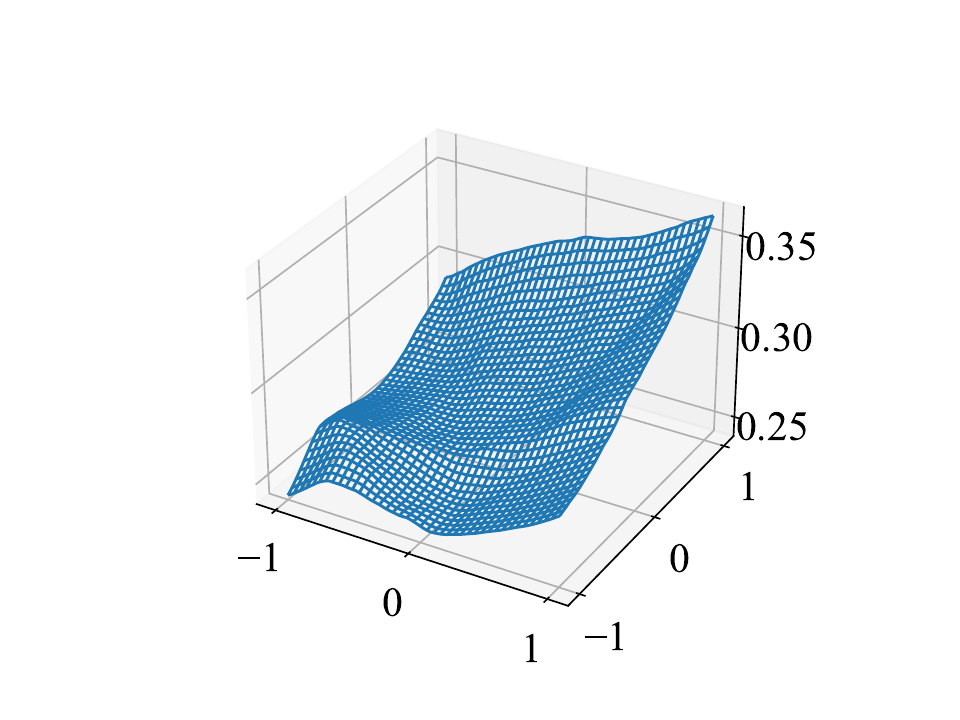}}
	\subfigure[Approximate Han-3. Left to right: the original Han-3, the trained FC-1, the trained Han-3.]{
		\includegraphics[width=.25\textwidth,trim=80 30 70 30, clip]{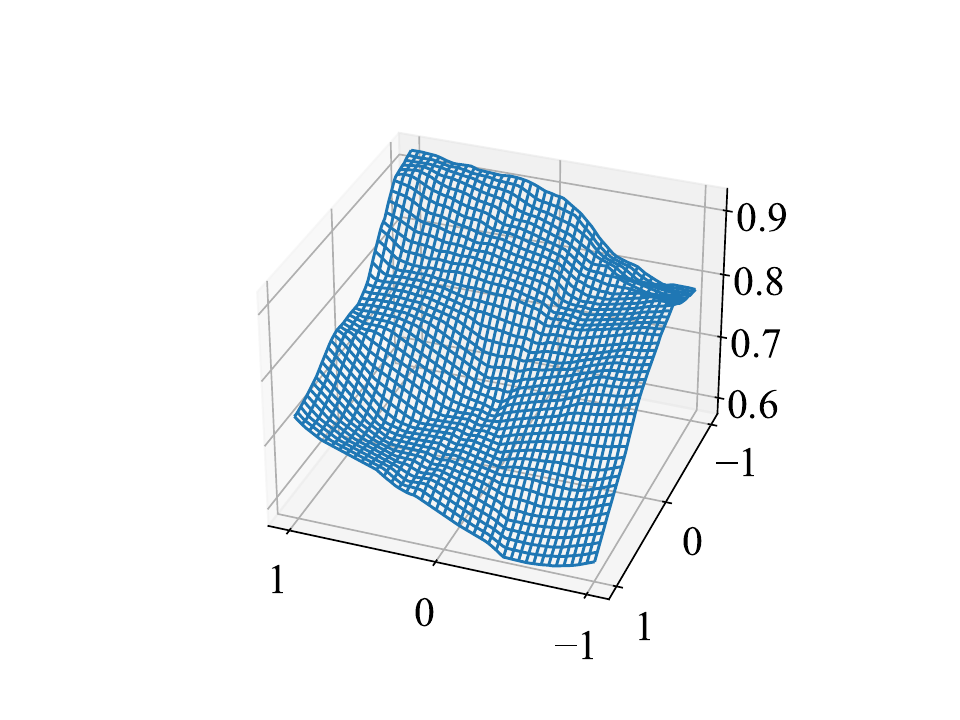}
		\includegraphics[width=.25\textwidth,trim=80 30 70 30, clip]{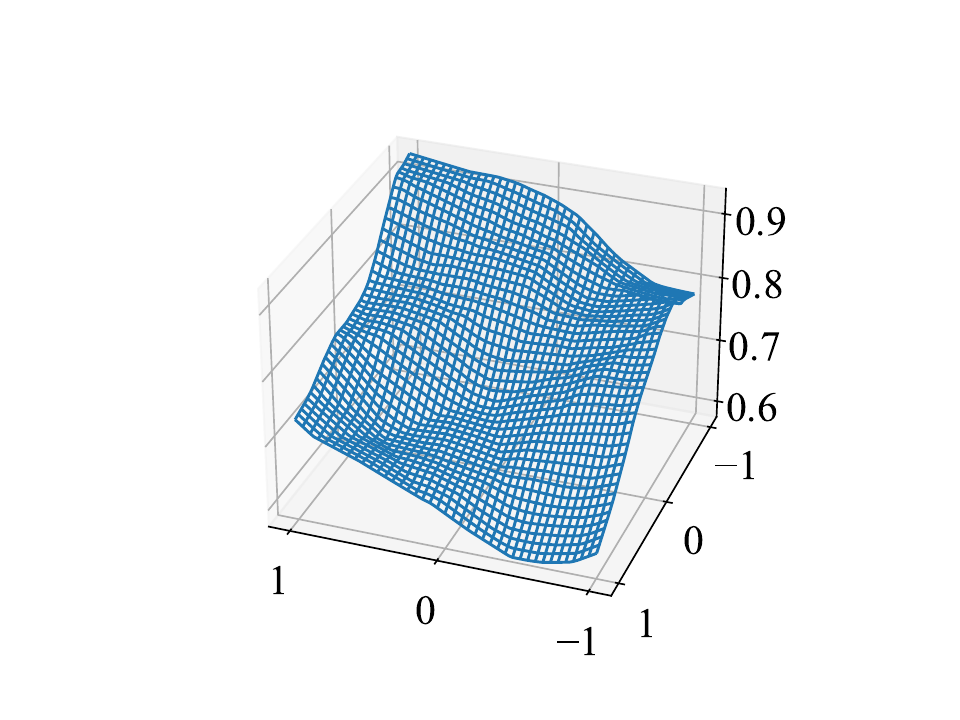}
		\includegraphics[width=.25\textwidth,trim=80 30 70 30, clip]{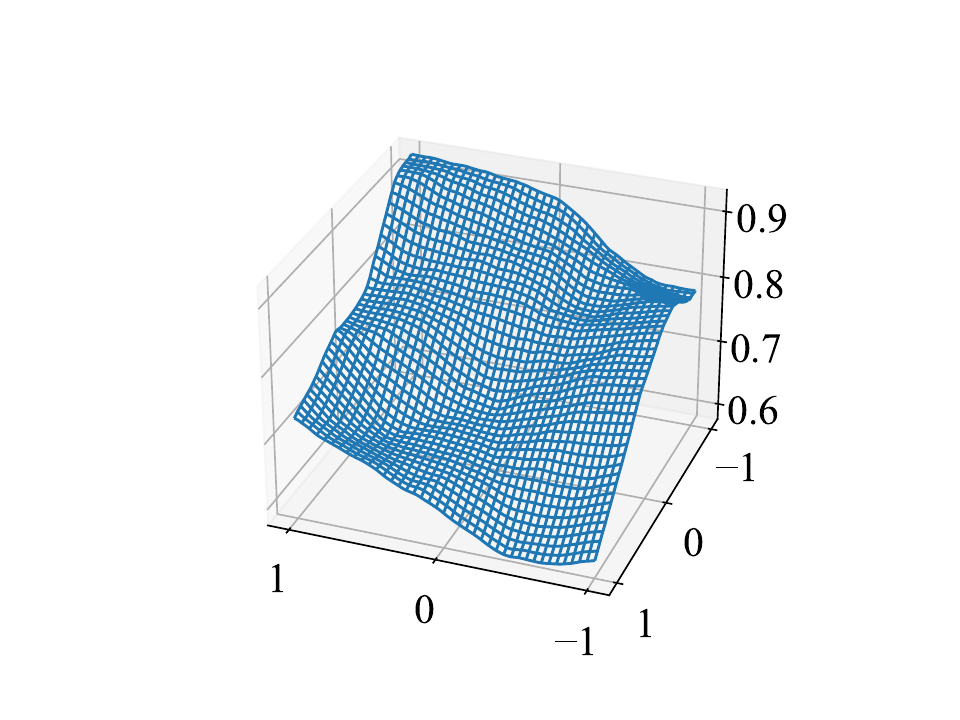}}
	\vspace{-.2cm}
	\caption{Landscape of $\|F_{FC}(x)\|_2$ and $\|F_{Han}(x)\|_2$ in one instance, where $d=50$.}
	\label{fig:approximation-result}
\end{figure*}

In a neural network with an element-wise activation function constructed from $\phi$ in \eqref{abs2}, the sign will be absorbed by weights and the constants will be absorbed into bias terms.  Therefore, without loss of generality \eqref{abs2} simply reduces to the absolute-value function.  We summarize the most relevant facts as a proposition below, where without confusion we use $\phi$ inter-changeably with either the scalar case or its element-wise extension in $\R^n$.
\begin{proposition}
	Let $\phi(x): \R^d\rightarrow\R^d$ be an element-wise activation function.  The Jacobian matrix of $\phi$ at $x \in\R^n$, which is diagonal if it exists, is orthogonal if and only $\phi$ satisfies the unit-derivative condition \eqref{unit-d}.  Moreover, among all activation functions with Property~A, the one with maximum differentiability is the absolute-value function $\phi(\cdot)=|\cdot|$.
\end{proposition}

\section{Properties of HanNet} \label{sec: property han}

We demonstrate that the Han-layer function exhibits a Lipschitz constant of 1 and possesses a stable gradient. Furthermore, we conduct an experimental evaluation to mutually approximate Han and FC models, which validates the feasibility of replacing the FC-layer with Han-layers.

\begin{proposition} \label{prop: 1-lip Han-layer}
	$\varphi(x;u,b) = |H(u)x+b| $ is a 1-Lipschitz function.
\end{proposition}
\begin{proof}
	\begin{align}
		&\||H(u)x_1+b|-|H(u)x_2+b|\|_2 \\
		\leq  &\|H(u)x_1-H(u)x_2\|_2 \nonumber \\
		\leq & \|H(u)\|_2\|x_1-x_2\|_2  \nonumber \\
		\leq & \|x_1-x_2\|_2
	\end{align}
	which completes the proof.
\end{proof}

\subsection{Gradient stability}\label{GL-stability}

Let the neural network function $F_L(x,\bW,\bb)$ be defined in \eqref{def:FL}. For HanNet, with the notation $\bu = \{u_k\}_{k=1}^L$ with all $u_k \ne 0$, the  $G_L$-matrix takes the form
\begin{equation}\label{def:hanG_L}
	G_L(x,\bu,\bb) = \prod_{k=1}^L H(u_k)\nabla\phi(z_k),
	\vspace{-.05cm}
\end{equation}
where $H(u_k)$ is defined by \eqref{eq:H}.
{Evidently, the diagonal matrices $\nabla\phi(z_k)$ are orthogonal matrices whose diagonal elements are equal to $\pm 1$ (with the convention $|\phi'(0)|=1$). Then the product $\prod_{k=1}^L H(u_k)\nabla\phi(z_k)$ is always orthogonal, which implies that the gradient will never vanish or explode.}  We state this fact as the following proposition.
\begin{proposition}\label{prop:hannet}
	The $G$-matrix for HanNet, that is, $G_L(x,\bu,\bb)$ defined in \eqref{def:hanG_L}, remains orthogonal for any $x$, any $\bu$ (with all $u_k \ne 0$), any $\bb$, and any integer $L>0$.
\end{proposition}

\subsection{Mutual Approximation: FC and Han}

We aim to empirically evaluate the efficacy of using Han-layers (of $O(d)$ complexity) to replace FC-layers (of $O(d)$ complexity), mainly to get an idea on how many Han-layers are needed to approximate one FC-layer. For this purpose, we define the following two models:
\begin{align*}
	&\textrm{FC-1} := \psi_{out} \circ [\psi_1] \circ \psi_{in} (x), \\
	&\textrm{Han-k} := \psi_{out} \circ [\varphi_{k} \circ... \circ\varphi_{1}] \circ \psi_{in}(x),
\end{align*}
where $\psi_{in}$ ($\psi_{out}$) denotes an input (output) layer in both models. Model FC-1 has one hidden FC-layer $\psi_1$, and Han-$k$ composes of $k$ Han-layers $\varphi$, and the width of all hidden layers in both models is fixed at $d$. FC-1 uses ReLU as the activation function, while Han-$k$ uses ABS.

Specifically, we experiment with $k=3$ and $d$ is from 30 to 100. The number of model parameters in FC-1 ranges from 900 to 10000 roughly, while for Han-3 it is from 180 to 600 roughly. We randomly sample 2000 points from the square $(-1,1)^2$ as the training set, and 10000 points from the square for testing. All FC models are initialized using random orthogonal initialization. We use the mean squared error (MSE) loss function.  Further experimental details are outlined in Table~\ref{tab: approximation settings}. In order to minimize the impact of training interventions on the final results, we explore a list of hyper-parameters to identify the best-performing model. Specifically, the initial learning rate is annealed to 0.2 for 6/10 and 9/10 of the training duration.

\begin{table}[ht]
	\caption{Settings in mutual approximation experiments. }
	\centering
	\resizebox{.9\columnwidth}{!}
	{
		\begin{tabular}{ccccc}
			\toprule[1.2pt]
			Optimizer & LR  &  weight decay & batch size & epochs \\ \midrule
			\{SGD, Adam\} & \{0.0001, 0.001, 0.01, 0.1\}  & \{0.0, 0.001\}  & 100 &  1000 \\
			\bottomrule[1.2pt]
		\end{tabular}
	}
	\label{tab: approximation settings}
\end{table}

\begin{figure}[ht]
	\centering
	\includegraphics[width=.4\textwidth,trim=50 20 10 10, clip]{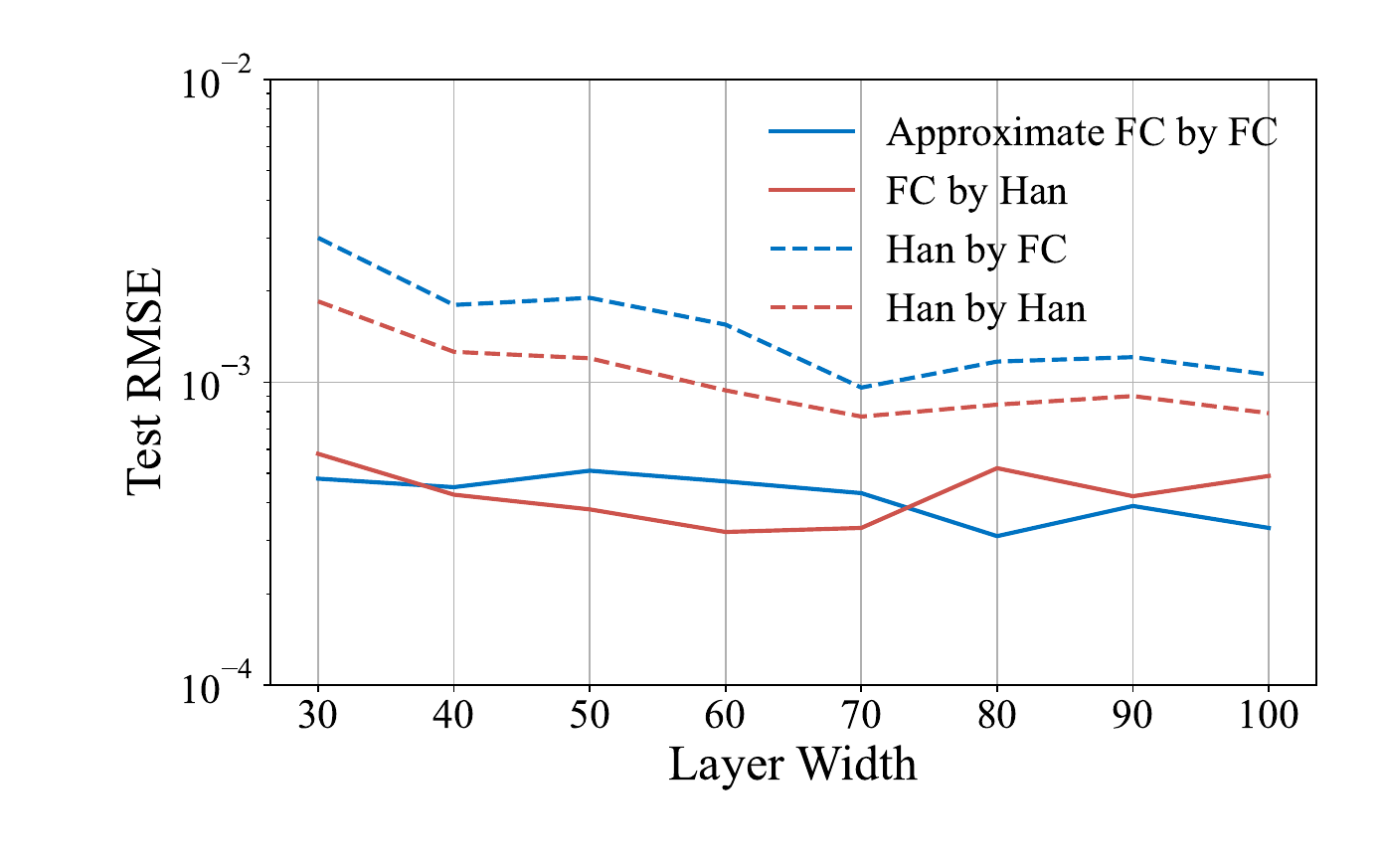}   
	\caption{\textbf{The average root mean squared error (RMSE) on 5 instances.} Blue line: FC-1 approximates itself, red line: Han-3 approximates FC-1, blue dash line: FC-1 approximates Han-3, red dash line: Han-3 approximates itself. }
	\label{fig:fcvshan-approx}
\end{figure}

The experimental results are presented in Figures~\ref{fig:approximation-result} and \ref{fig:fcvshan-approx}. We observe that the test errors of the two models fitting each other are around $10^{-3}$ or slightly below. The obtained landscapes are consistent with the original models, suggesting the potential for replacing FC-layers with a small number of Han-layers that is far fewer than width $d$.

\begin{table}[ht]
	\caption{%
	{The test RMSE for input dimensions ranging from 2 to 4, where the width $d$ in all hidden layers is $d=100$. } }
	\label{tab:dim2to4}
	\centering
	\resizebox{.8\columnwidth}{!}
	{
		\begin{tabular}{l|cc|cc}
			\toprule[1.2pt]
			& \multicolumn{2}{c|}{approximate FC-1} & \multicolumn{2}{c}{approximate Han-3} \\
			trained model & FC-1 & Han-3 & FC-1 & Han-3 \\
			\midrule
			input dim 2 & 0.00033 & 0.00049 & 0.00107 & 0.00079 \\
			input dim 3 & 0.00086 & 0.00078 & 0.00251 & 0.00237 \\
			input dim 4 & 0.00184 & 0.00192 & 0.00478 & 0.00494 \\
			\bottomrule[1.2pt]
		\end{tabular}
	}
\end{table}

Moreover, we extend the input and output dimensions to 4 while maintaining $d=100$, and the results are summarized in Table~\ref{tab:dim2to4}.  Our findings reveal that as the input data dimensionality increases, the RMSE values for both models also increase. However, the mutual approximation performance of the two models remains comparable. Overall, the above experiment confirms the ability of HanNet and FCNet to approximate each other, thereby shedding light on the possibility of reducing the number of model parameters.

\section{Stylized Datasets}\label{result:Checkerboard}	

\begin{figure}[ht]
	\centering
	\includegraphics[width=.16\textwidth,trim=110 40 100 60, clip]{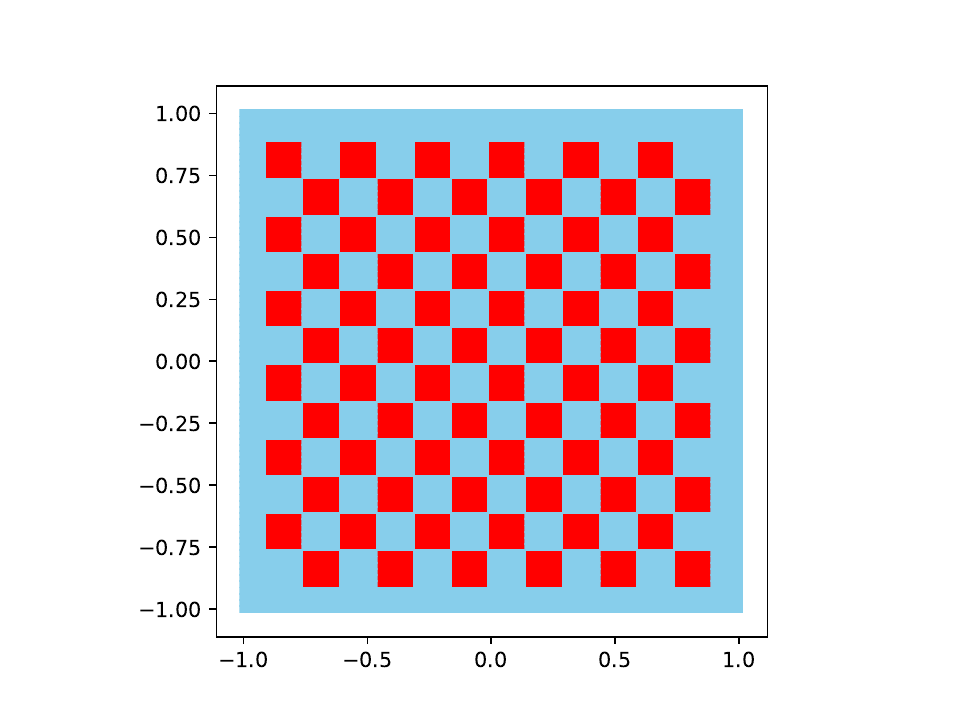}
	\includegraphics[width=.16\textwidth,trim=110 40 100 60, clip]{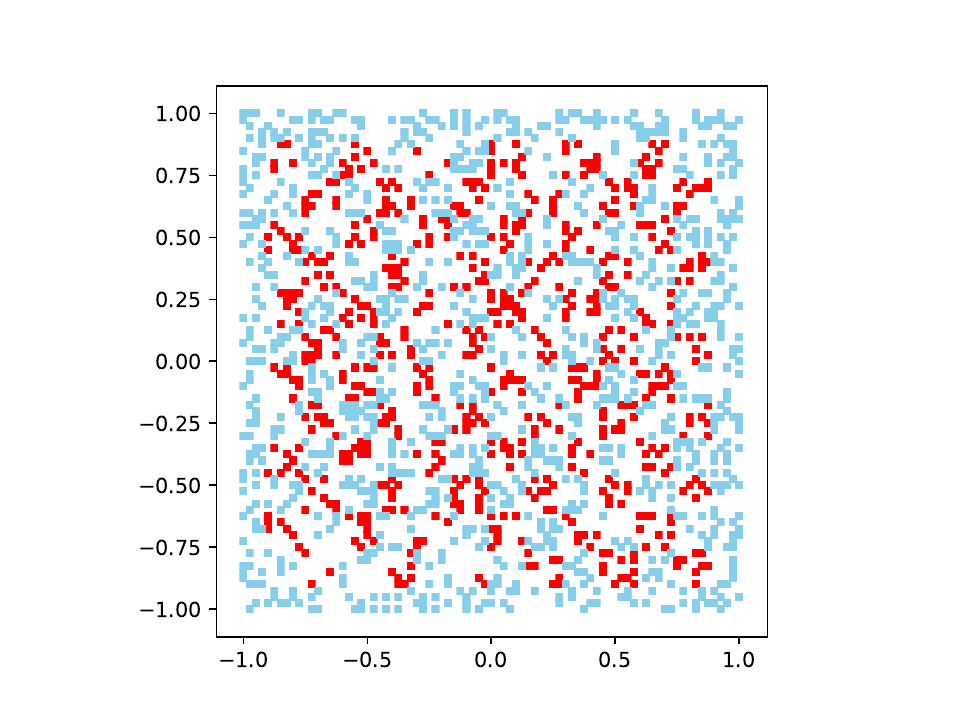}
	\caption{Checkerboard datasets. In the right figure, the dots represent the training set~(25\%). }
	\label{fig:checkerboard data}
\end{figure}

In this section, we first evaluate our approach on a synthetic checkerboard dataset, depicted in Figure~\ref{fig:checkerboard data}, which consists of 6561 mesh points over the square $[-1, 1]^2$. Our experiments confirm that achieving a perfect solution on this dataset using standard MLPs is extremely challenging.

We adopt MSE loss and select SGD optimizer with a momentum of 0.9. Our model is trained using a batch size of 100 and 40000 iterations. To solve each test instance, we perform SGD with 10 distinct initial learning rates:
\[
\{0.001, 0.005, 0.01, 0.025, 0.05, 0.075, 0.1, 0.25, 0.5, 1\}
\]
and select the best result as the output. Moreover, the initial learning rate is annealed by a factor of 5 at the fractions 1/2, 7/10, and 9/10 of the training durations. 

Our experiments show HanNets possess an unusually high level of generalization ability in Figure~\ref{fig:checkerboard}. A HanNet outperforms FCNets significantly and produces a nearly perfect result.

\begin{figure}[ht]
	\centering
	\subfigure[FCNet: the best test accuracy is about 87\%.]{
		\includegraphics[width=.4\textwidth]{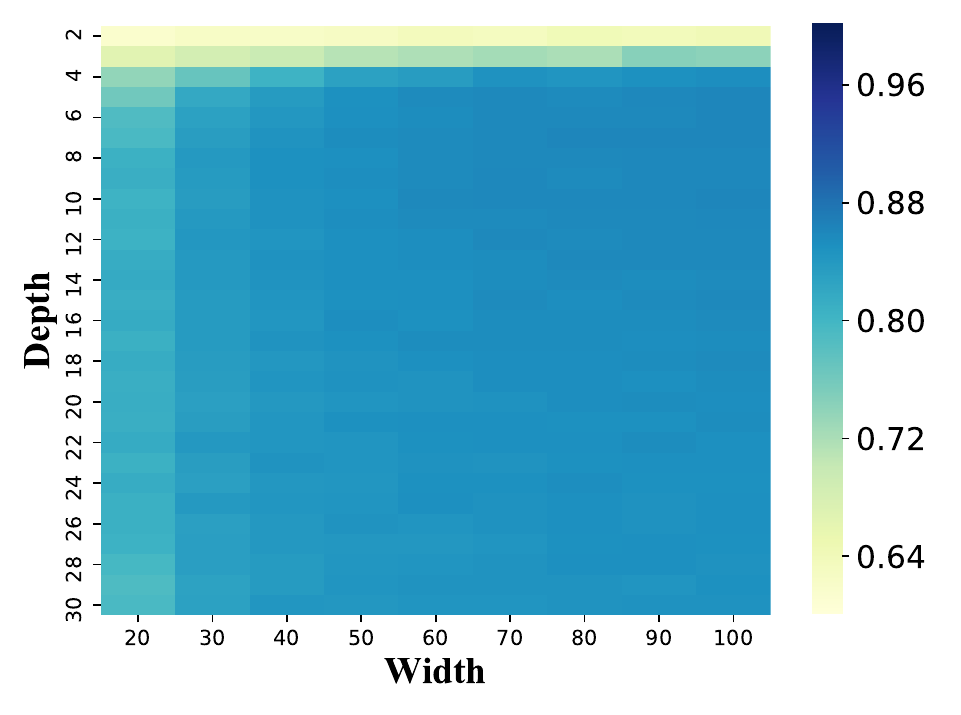}}
	\subfigure[ HanNet: the best test accuracy is over 99\%.]{
		\includegraphics[width=.4\textwidth]{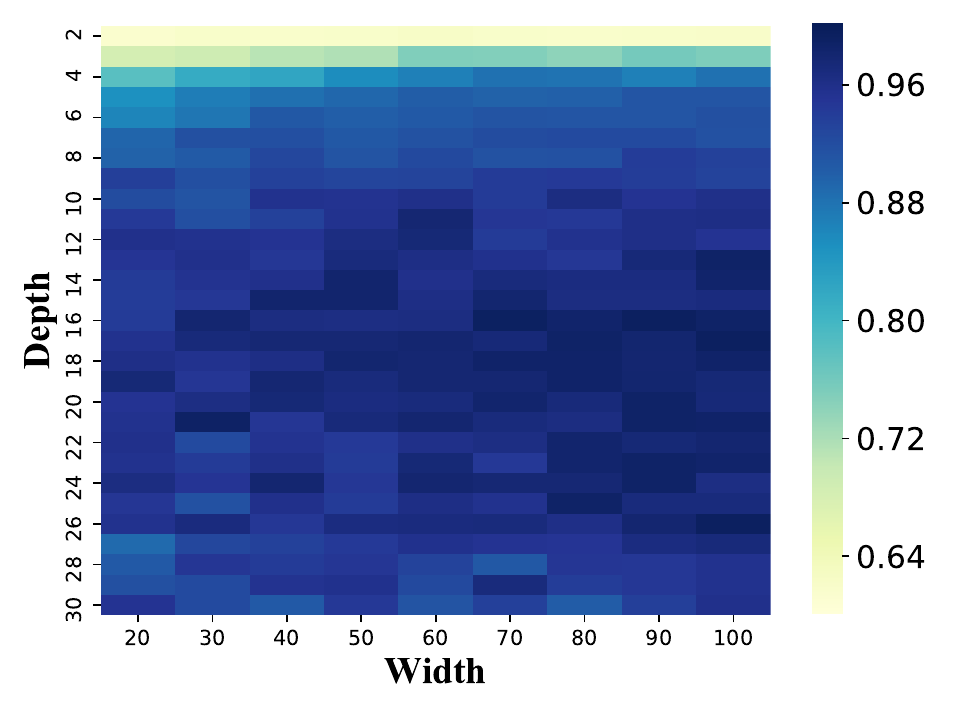}}
	\caption{Heat maps of FCNets and HanNets. }
	
	\label{fig:checkboard-heatmap}
\end{figure}

We also compare various FCNets with HanNets (FCNets with skip connection were also compared, but did not yield better results than the FCNets). We generate more than 200 pairs of FCNets and HanNets, with widths ranging from 20 to 100 in increments of 10 and depths from 2 to 30. The test results are presented in Figure~\ref{fig:checkboard-heatmap} in the form of a heat map.

The results in Figure~\ref{fig:checkboard-heatmap} show that there is a significant gap in test accuracy between FCNets (around 85\%) and HanNets (over 99\%) for a wide range of test cases. The remarkable near-perfect results achieved by HanNets, without any explicit regularization, are both impressive and stable, as similar results can be obtained from other random 25\% vs. 75\% data splits. We also observe that over-parameterization is not a necessary condition for achieving near-zero test error, as demonstrated in one of the cases in Figure~\ref{fig:checkboard-heatmap} (Bottom) where 400 parameters are sufficient. 

\subsection{A preliminary explanation }

To partially explain HanNet's exceptional performance on the checkerboard dataset, we give an intuitive viewpoint based on the amount of nonlinearity provided by different models. In most feed-forward networks, the main source of nonlinearity is from the applications of nonlinear activation functions.  Given an MLP-layer that maps from $\mathbb{R}^d$ to $\mathbb{R}^d$, the number of activation-function applications is $d$, with $d^2$ parameters in the associated weight matrix $W$.  Consequently, the activation-function utilization rate is $1/d$ (i.e., $d$ versus $d^2$), which serves as a measure of nonlinearity normalized with respect to the parameter count.  In contrast, with $d$ parameters in $u$, our Han-layer has an activation-function utilization rate 1 (i.e., $d$ versus $d$).  Therefore, several Han-layers together can potentially produce more nonlinearity than a single MLP layer can, while still using far fewer parameters.   Moreover, as we discussed in Section~\ref{GL-stability}, the absolute-value activation, coupled with Householder weighting, gives gradient stability. 

It is reasonable to argue that, with high nonlinearity and gradient stability, a HanNet would be better equipped than an FCNet to approximate complex shapes as in the checkerboard dataset and capture the underlying patterns.  On the other hand, an FCNet, characterized by over-parameterization, may be able to readily fit training data by memorizing them, but still fail to fully discern the underlying patterns.  This seems to explain, at least partially, the observed generalization discrepancy in Figure~\ref{fig:checkerboard}.

\subsection{Ablation Study on Checkerboard} \label{sec: ablation}

We conducted an ablation study on a $100\times 20$ network framework with seven different configurations to investigate how much Householder weighting and ABS activating contribute to the remarkable results on the checkerboard dataset. Table~\ref{tab: ablation} lists the results, where we used either Householder (HH) or fully connected (FC) layers with/without orthogonal regularization, and either ABS, ReLU, MaxMin or GroupSort~\cite{groupsort} activation functions. The latter two can keep the gradient norm unchanged~\cite{groupsort}. We obtained orthogonal FCNets~(OrthFC) by regularizing each weight matrix $W$ as~\cite{brock2017neural}: 
\[
\lambda \|W^{\T} W-I\|_F^2,
\]
where $\lambda=0.1$ in our experiments.

Table~\ref{tab: ablation} suggests that Householder weighting and ABS activating are equally critical to the 99\% generalization accuracy of the HanNet. Other activation functions influence the test performance badly as that happens on the Householder Layer. Furthermore, OrthFC cannot achieve the same accuracy as HanNet, although each of its weight matrices is orthogonal.
\begin{table}[ht]
	\caption{\textrm{Ablation study on a $100 \times 20$ network framework}: Effects of layer and activation types on performance. HH denotes Householder and FC denotes fully connected layers. }
	\centering
	\resizebox{1.\columnwidth}{!}{
		\begin{tabular}{c|ccc|cccc|c}
			\toprule[1.2pt]
			&\multicolumn{3}{c}{Layer type} & \multicolumn{4}{c}{Activation type} & \multirow{2}{*}{Test accuracy}\\
			& HH & FC & OrthFC & ABS & ReLU & MaxMin & GroupSort~(5)  &  \\
			\midrule
			\textbf{(a)}  & \Checkmark & -- & -- & \Checkmark  & -- & -- & -- & 99.2\%\\
			\textbf{(b)}  & \Checkmark & -- & --   & --  &  \Checkmark  & --   & --   & 66.2\% \\
			\textbf{(c)}  & \Checkmark & -- & --   & --  & --  &  \Checkmark    & --   & 77.6\% \\
			\textbf{(d)}  & \Checkmark & -- & --   & --    & --   & --   &  \Checkmark & 81.4\% \\
			\textbf{(e)}  & -- & \Checkmark & --  & \Checkmark &  -- &  -- &  --  &  85.3\% \\
			\textbf{(f1)}  & -- & -- & \Checkmark   & \Checkmark &  -- &  -- &  --  &  86.7\% \\
			\textbf{(f2)}  & -- & -- & \Checkmark  &  -- & \Checkmark  &  -- &  --  &  75.5\% \\
			\bottomrule[1.2pt]
		\end{tabular}
	}
	\label{tab: ablation}
\end{table}

\subsection{Another Stylized Dataset} 

The remarkable performance of HanNet on the checkerboard dataset is not an isolated case, as shown by the experiment in~\cite{hong2022activation}. Here, we fit another function $g(x)$ using HanNet and FCNet, defined by the sum of two sine-products (of low and high frequencies)
\[ g(x) = \sin(2\pi x_1) \sin(2\pi x_2) + \sin(10\pi x_1) \sin(10 \pi x_2),\]
where $x = (x_1, x_2) \in [0, 1]^2$ is evaluated on the same grid as in the checkerboard dataset (also 25\% of the data used for training). All models have a width of 200 and a depth of 20, and are trained with the same SGD optimizer and annealing strategy on checkerboard for 80000 iterations.
\begin{figure}[ht]
	\centering
	\includegraphics[width=.23\columnwidth,trim=110 40 100 60, clip]{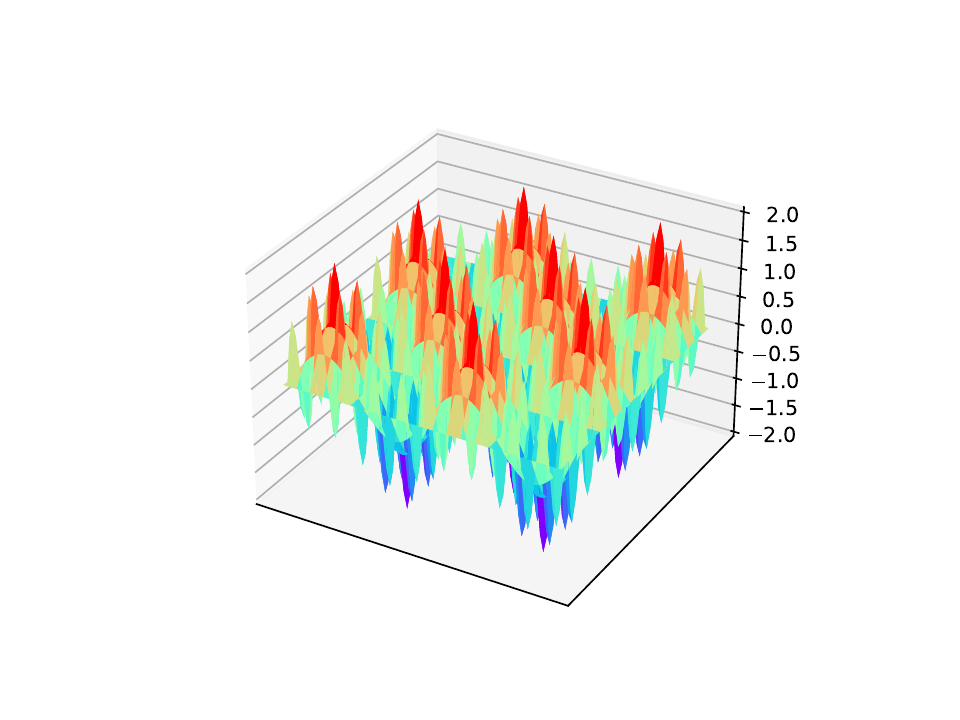}
	\includegraphics[width=.23\columnwidth,trim=110 40 100 60, clip]{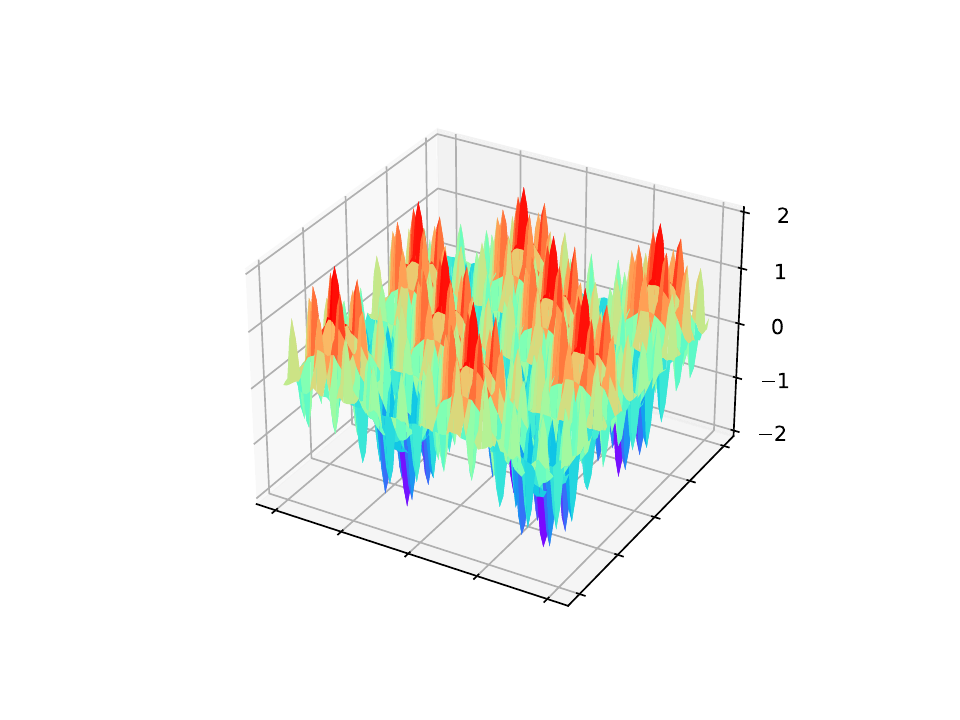}
	\includegraphics[width=.23\columnwidth,trim=110 40 100 60, clip]{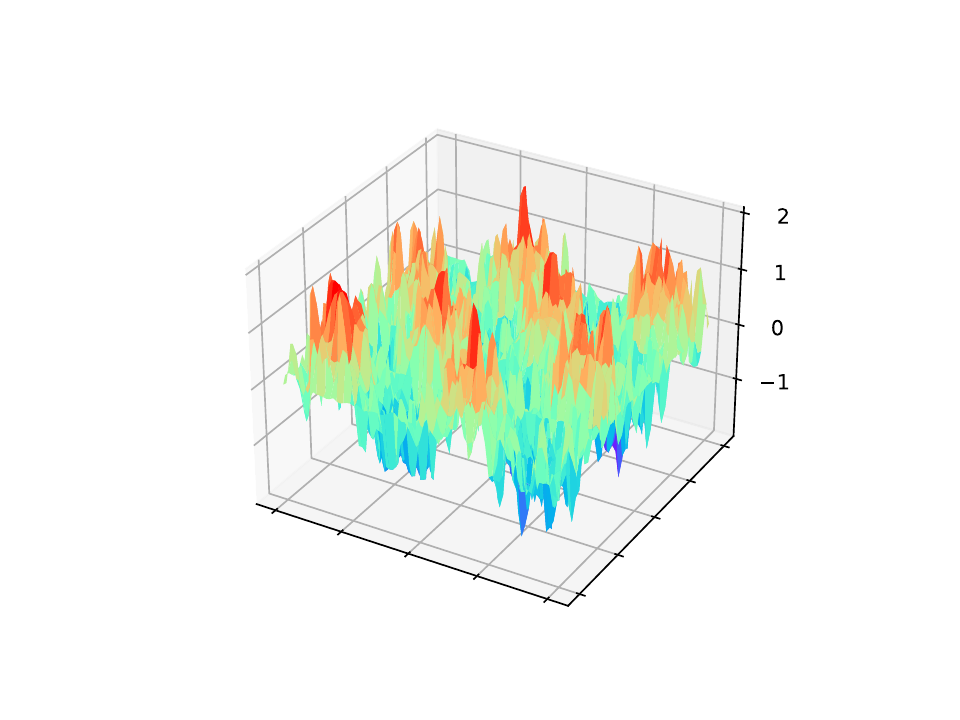}
	\includegraphics[width=.23\columnwidth,trim=110 40 100 60, clip]{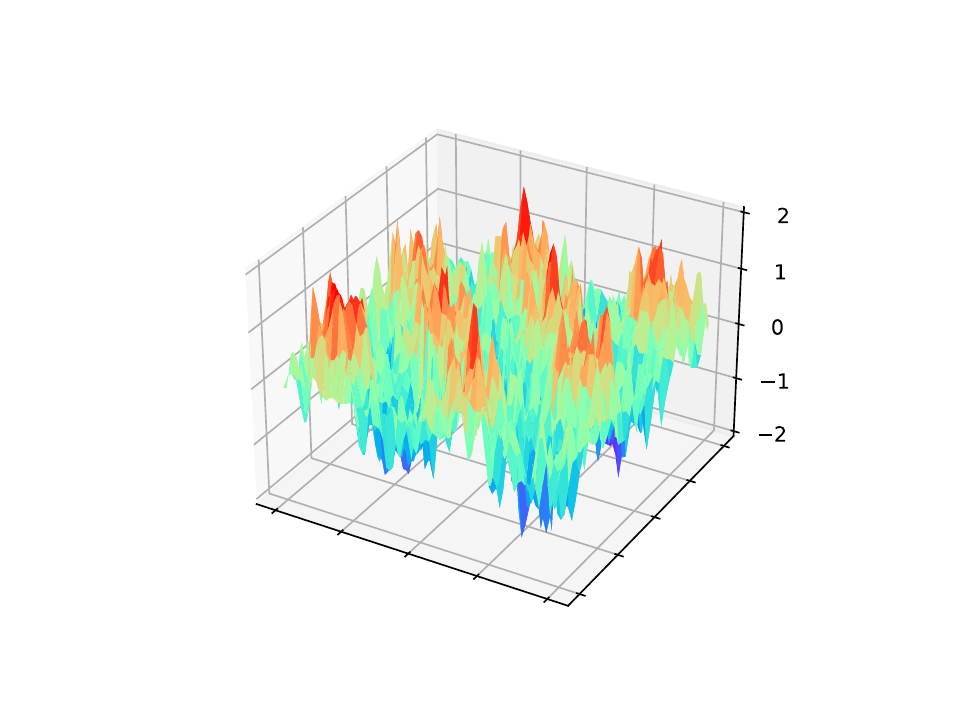}

	\includegraphics[width=.23\columnwidth,trim=110 50 100 60, clip]{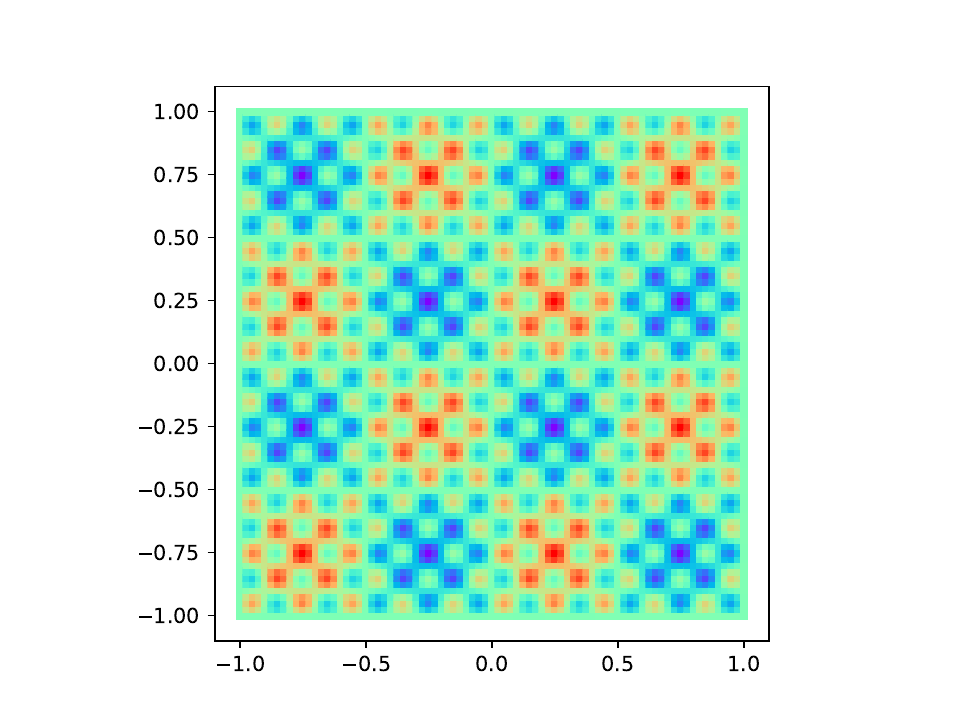}
	\includegraphics[width=.23\columnwidth,trim=110 50 100 60, clip]{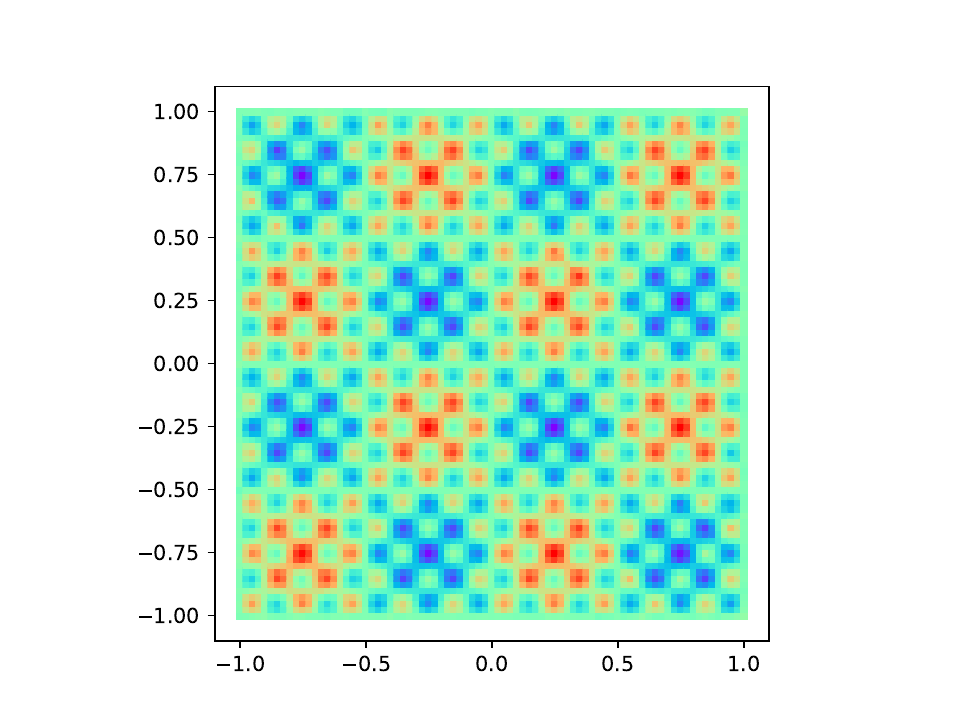}
	\includegraphics[width=.23\columnwidth,trim=110 50 100 60, clip]{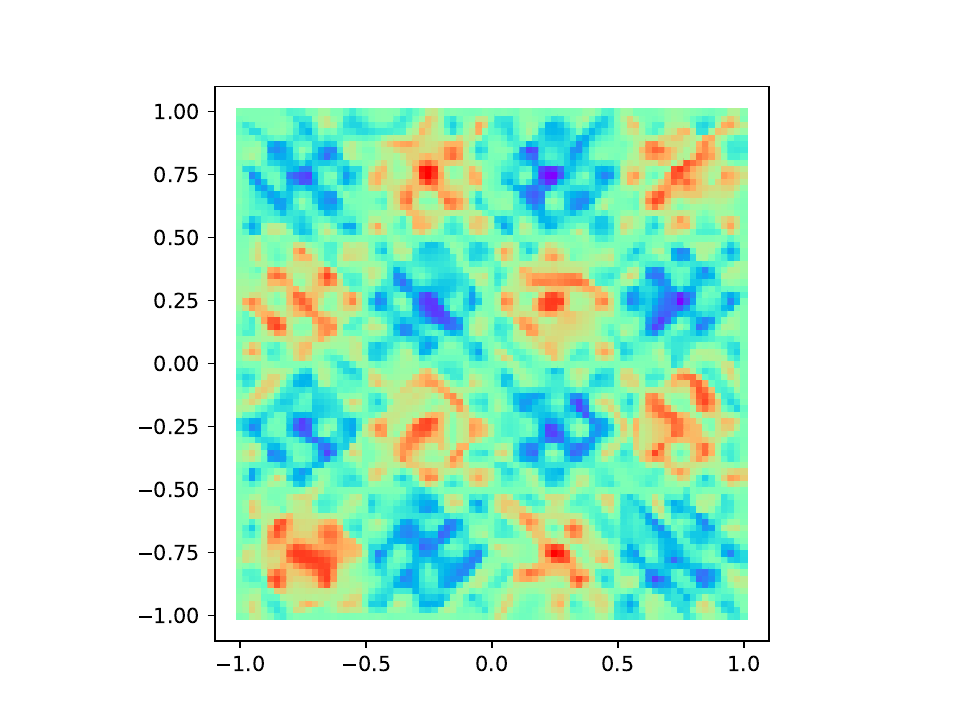}	
	\includegraphics[width=.23\columnwidth,trim=110 50 100 60, clip]{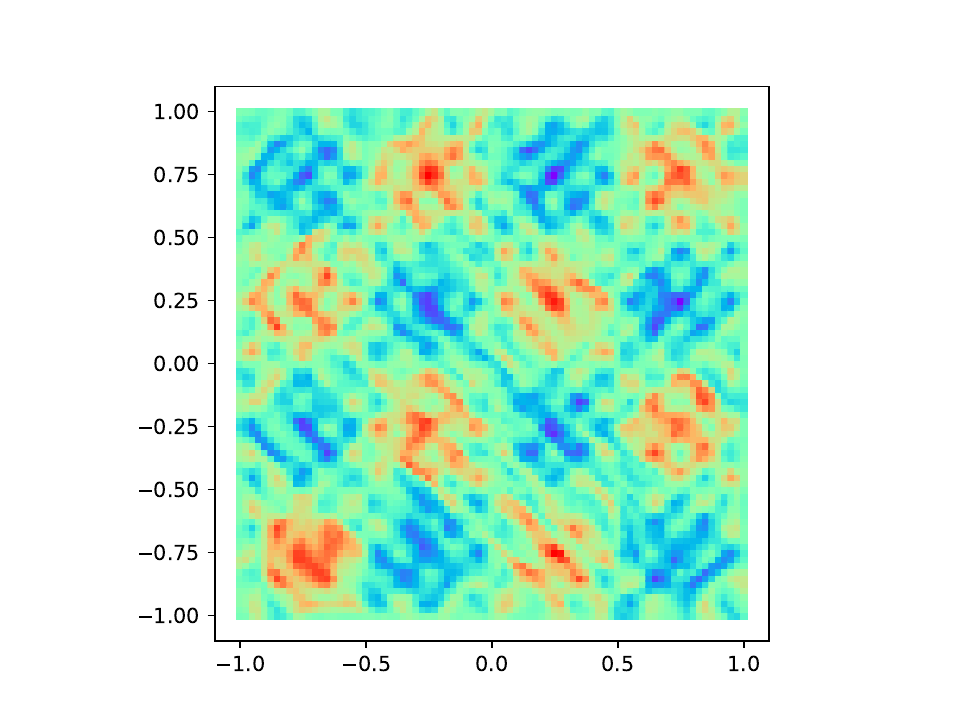}

	\caption{ Landscapes and top views for FC and Han models. From left to right: the original data, HanNet, FC-ReLU, FC-ABS. The test loss by HanNet is 0.034 while it is about 0.37 by the other two models.}
	\label{fig: experiment on fourier}
\end{figure}

The results in Figure~\ref{fig: experiment on fourier} demonstrate a clearly superior performance of HanNet over FCNet, indicating that the advantage of HanNet is not limited to the checkerboard dataset only.  Of course, further study is needed to fully understand this advantage.

\section{Experiments on Larger Datasets}

We present extensive numerical experiments to convincingly demonstrate the efficacy of the Han model, using the following datasets.

We conduct an investigation into the impact of the Han model on a widely used MNIST dataset~\cite{lecun1998gradient} and explored its robustness. We also evaluate the regression performance on five classic datasets, namely Bikesharing, Cal Housing, Elevators, Parkinsons, and Skillcraft~\cite{fanaee2014event, pace1997sparse, Dua:2019, little2007exploiting, thompson2013video}. In addition, we assess the performance of the proposed model on four widely used image classification datasets, namely CIFAR10, CIFAR100, STL10, and Downsampled-ImageNet~\cite{krizhevsky2009learning, coates2011analysis, chrabaszcz2017downsampled}, where the last one is a downsampled version of ImageNet for affordability reasons.

All training is performed using PyTorch~\cite{paszke2019pytorch} on a shared cluster. Multiple runs are performed for each test instance, starting with different random initializations of the model parameters. All reported values are the average of at least five runs.

\subsection{MNIST Dataset} 

\subsubsection{Same performance, fewer parameters}

Our study aims to evaluate the model's performance on MNIST dataset~\cite{lecun1998gradient}. In this study, we consider one hidden layer FCNet model,  and $k$ Han-layer models, where $k$ ranges from 2 to 20. Notably, the width of each layer is fixed to the input size ($784$) in two models. To train our models, we use 10,000 instances for training and the remaining instances for testing. The details of other hyperparameters used in this study are presented in Table~\ref{tab:settings-in-general}. We anneal the initial learning rate to 0.2 during half of the training period.
\begin{table}[ht]
	\caption{Settings on MNIST. }
	\centering
	\resizebox{.9\columnwidth}{!}
	{	
		\begin{tabular}{lccccc}
			\toprule[1.2pt]
			Optimizer & LR   & loss   & weight decay &batch size & epochs \\ \midrule
			Adam & 0.001    & cross entropy & 0.001 & 256 &  100 \\
			\bottomrule[1.2pt]  
		\end{tabular}
	}
	\label{tab:settings-in-general}
\end{table}
\begin{figure}[ht]
	\centering
	\includegraphics[width=.36\textwidth,trim=0 0 0 0, clip]{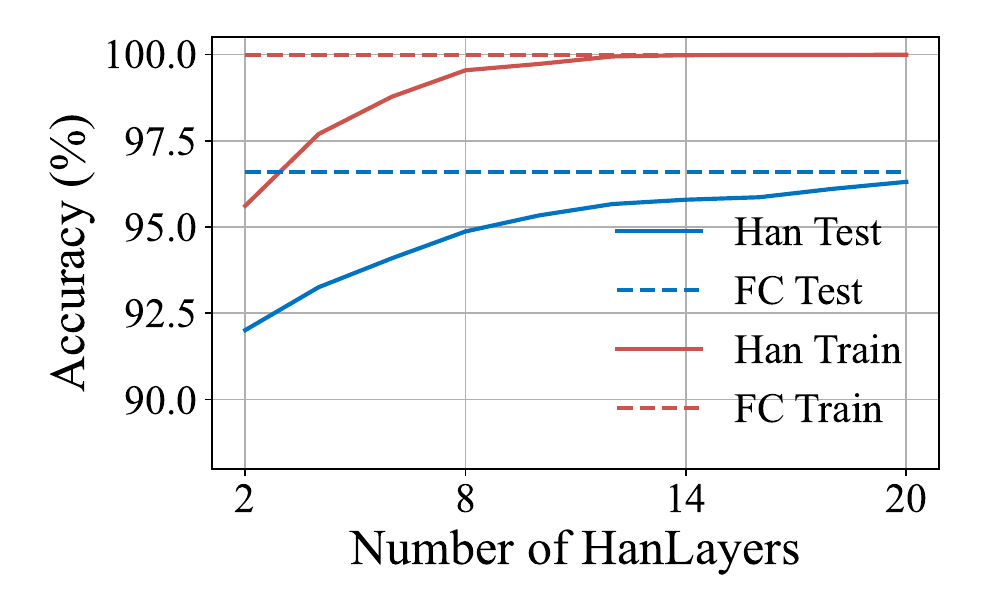}  
	\caption{The training (red) and testing (blue) performances in one hidden-layer FCNet and  $k$-layer HanNets~($k$ from 1 to 20). }
	\label{fig:fcvshan generalization}
\end{figure}

Figure~\ref{fig:fcvshan generalization} depicts  that a 20-layer HanNet achieves the same train/test accuracy as FCNet on the clean dataset, while utilizing approximately $40K$ and $600K$ parameters, respectively.

\subsubsection{Robustness by 1-Lipschitz}

Many studies have established the robustness of 1-Lipschitz models against adversarial examples, for instance as evidenced by~\cite{tsuzuku2018lipschitz}. We investigate the contrast in robustness between Han and FC models using identical settings as in the previous subsection. The baseline models are one hidden layer FCNet with a width of 784, and a 20 hidden Han-layer model with the same width. 

Adversarial examples are generated using PGD and CW attacks as detailed in~\cite{carlini2017towards, madry2018towards} and summarized in Table~\ref{tab: attacks}. We extend our evaluation to defensive FC models trained with adversarial training and TRADES~\cite{madry2018towards, zhang2019theoretically} using $\ell_\infty$ FGSM attack samples with a maximum perturbation of $\rho=0.1$, in addition to vanilla FCNet. To improve the robustness of HanNet, we adjust the value of weight decay in the optimizer and select the value from $\{0.01,0.05,0.1,0.2\}$ that offers the best robustness while maintaining clean accuracy above 0.9.
\begin{table}[htp]
	\centering
	\caption{Attack settings. $\rho$ is the maximum perturbation, $\alpha$ is the magnitude of the perturbation in each iteration, $c$ is the regularization constant value.}
	\begin{tabular}{ll}
		\toprule[1.2pt]
		PGD $\ell_\infty$ &  30 steps , $\rho = 0.1,0.2,0.3$, $\alpha = 0.01 $\\
		PGD $\ell_2$ &  30 steps , $\rho = 1.0,2.0,3.0$, $\alpha = 0.1 $ \\
		CW & 1000 steps, $c=1,3,5$\\
		\bottomrule[1.2pt]
	\end{tabular}
	\label{tab: attacks}
\end{table}

Table~\ref{tab:robustness-on-mnist} presents empirical evidence that suggests vanilla HanNet surpasses vanilla FCNet in robustness under attack, despite having a clean accuracy that is 0.04 lower than FCNet. Moreover, HanNet's performance is comparable with that of defensive FC models and in certain instances, outperforms them. For instance, under CW attack with $c=5$, HanNet yields approximately 0.65 accuracy, while FCNets only achieve about 0.2 accuracy. The findings of our study demonstrate that HanNet can achieve robustness on par with defensive FCNet on the MNIST dataset under some attack situations.
\begin{table}[htp]
	\centering
	\caption{The robust accuracy with respect to CW and PGD attacks. VT and AT denote vanilla training and adversarial training respectively.}
	\resizebox{.9\columnwidth}{!}{
	\begin{tabular}{ll|c|ccc}
		\toprule[1.2pt]
		\multicolumn{1}{l}{{  }} & {  } &   HanNet & \multicolumn{3}{c}{{  FCNet}} \\
		\multicolumn{1}{l}{{  }} & {  } & {  VT} & {  VT} & {  Trades} & {  AT} \\
		\midrule
		\multicolumn{1}{l}{{  clean}} & {  } & {  0.925} & {  0.964} & {  \textbf{0.969}} & {  0.967} \\
		\midrule
		{  } & {  c=1} & {  \textbf{0.887}} & {  0.684} & {  0.828} & {  0.815} \\
		{  } & {  c=3} & {  \textbf{0.757}} & {  0.069} & {  0.437} & {  0.348} \\
		\multirow{-3}{*}{{  CW}} & {  c=5} & {  \textbf{0.653}} & {  0.032} & {  0.216} & {  0.176} \\
		\midrule
		{  } & {  $\rho=1$} & {  0.783} & {  0.705} & {  \textbf{0.866}} & {  0.859} \\
		{  } & {  $\rho=2$} & {  \textbf{0.529}} & {  0.121} & {  0.448} & {  0.412} \\
		\multirow{-3}{*}{{  PGD $\ell_2$}} & {  $\rho=3$} & {  \textbf{0.260}} & {  0.007} & {  0.071} & {  0.051}\\
		\midrule
		{  } & {  $\rho=0.1$} & {  0.753} & {  0.467} & {  0.861} & {  \textbf{0.867}} \\
		{  } & {  $\rho=0.2$} & {  0.406} & {  0.018} & {  \textbf{0.475}} & {  0.420} \\
		\multirow{-3}{*}{{  PGD $\ell_\infty$}} & {  $\rho=0.3$} & {  \textbf{0.075}} & {  0.001} & {  0.042} & {  0.041} \\
		\bottomrule[1.2pt]
	\end{tabular}
	}
	\label{tab:robustness-on-mnist}
\end{table}

\subsection{Regression}

We conduct experiments on these five real-world regression datasets in Table~\ref{tab: dataset}. We choose Adam~\cite{kingma2015adam} optimizer which seems to be the method of choice for several works in that area including~\cite{tsang2018neural} in Table~\ref{tab:adam regression}.  
\begin{table}[htp]
	\caption{Dataset statistics: $N$ is the number of samples, and Dim is the dimension of data vectors.}
	\centering
	\resizebox{.7\columnwidth}{!}{
	\begin{tabular}{llcc}
		\toprule[1.2pt]
		& Datasets & Dim & $N$ \\
		\midrule
		\multirow{5}{*}{Regression} &Bikesharing  & 15 & 17379\\
		&Cal Housing  & 8 &  20640 \\
		&Elevators  & 18 &  16599 \\
		&Parkinsons  & 20 &  5875 \\
		&Skillcraft  & 19 &  3338 \\
		\bottomrule[1.2pt]
		\vspace{0mm}
	\end{tabular}
	}
	\label{tab: dataset}
	\vspace{-.5cm}
\end{table}
\begin{table}[hbt]
	\centering
	\caption{Adam parameters on Regression Datasets.}
	\begin{tabular}{lcccc}
		\toprule[1.2pt]
		LR  &  weight decay &batch size & epochs \\ \midrule
		0.001  & 0.0  & 100 &  300 \\
		\bottomrule[1.2pt]  
	\end{tabular}
	\label{tab:adam regression}
\end{table}
We compare the performance of a HanNet with two FCNets.  Table~\ref{tab:regression} lists the relevant statistics for the 3 DNNs where depth refers to the number of hidden layers (there exist additional, data-size-dependent input/output layers). We see that in terms of parameter numbers FCNet1 and HanNet are comparable peers, while FCNet2 has about 15 times more parameters. 
\begin{table*}[ht]
	\caption{RMSE and R-squared on regression datasets. Lower RMSE and higher R-squared mean better model performance.  }
	\centering
	\resizebox{1.8\columnwidth}{!}{
		\begin{tabular}{lc|ccc|ccc}
			\toprule[1.2pt]
			&        & \multicolumn{3}{c|}{$\delta=0.8$} &\multicolumn{3}{c}{$\delta=0.2$}   \\ \midrule
			&      & HanNet     & FCNet1    & FCNet2   & HanNet   & FCNet1  & FCNet2    \\
			& Depth $\times$ Width (\#param) & 20 $\times$ 200 (10.6K)  & 5 $\times$ 50 (10.9K) & 5 $\times$ 200 (165K) & 20 $\times$ 200 (10.6K) & 5 $\times$ 50 (10.9K) & 5 $\times$ 200 (165K)  \\ \midrule
			\multirow{5}{*}{RMSE}   & Bikesharing  & \textbf{0.218}   & 0.241   & 0.223     & \textbf{0.311}  & 0.498   & 0.344   \\
			& Calhousing           & \textbf{0.431}          & 0.462                 & 0.459                  & \textbf{0.476}             & 0.506                 & 0.498                       \\
			& Elevators                  & \textbf{0.087}         & 0.090                  & 0.091               & \textbf{0.103}           & 0.140                  & 0.106                       \\
			& Parkinsons                & \textbf{1.235}        & 2.035                 & 1.399                  & \textbf{3.072}                & 5.034                 & 4.187                 \\
			& Skillcraft                      & \textbf{0.255}          & 0.282                 & 0.266                 & \textbf{0.276}          & 0.328                 & 0.301                \\ \midrule
			\multirow{5}{*}{R-squared} & Bikesharing             & \textbf{0.951}        & 0.938                 & 0.949           & \textbf{0.898}               & 0.744                 & 0.877                           \\
			& Calhousing                    & \textbf{0.813}       & 0.786                 & 0.788                  & \textbf{0.772}          & 0.743                 & 0.75                     \\
			& Elevators         & \textbf{0.878}               & 0.869                 & 0.869                  & \textbf{0.832}            & 0.671                 & 0.823                      \\
			& Parkinsons              & \textbf{0.986}         & 0.963                 & 0.982                    & \textbf{0.917}         & 0.777                 & 0.845                       \\
			& Skillcraft             & \textbf{0.546}           & 0.455                 & 0.507                 & \textbf{0.477}        & 0.266                 & 0.383                          \\ 
			\bottomrule[1.2pt]
	\end{tabular}}
	\label{tab:regression}
\end{table*}

We present test RMSE loss values and R-squared values in Table~\ref{tab:regression}, where all values are averaged over 5 trials. Clearly, HanNet outperforms its ``peer'' FCNet1 by a notable margin, especially when fewer training samples are used, and is also better than FCNet2 which uses 15 times more parameters. We mention that the best test performance of HanNet is statistically the same as~(or better than) that reported in~\cite{tsang2018neural} with an FCNet \& NIT model that uses more than $800K$ parameters. Furthermore, with fewer parameters, HanNet appears less influenced by overfitting, as seen in Figure~\ref{fig:elevators}. Figures for all datasets in the three models are provided in Appendix.
\begin{figure}[ht]
	\centering
	\includegraphics[width=.22\textwidth,trim=20 20 20 20, clip]{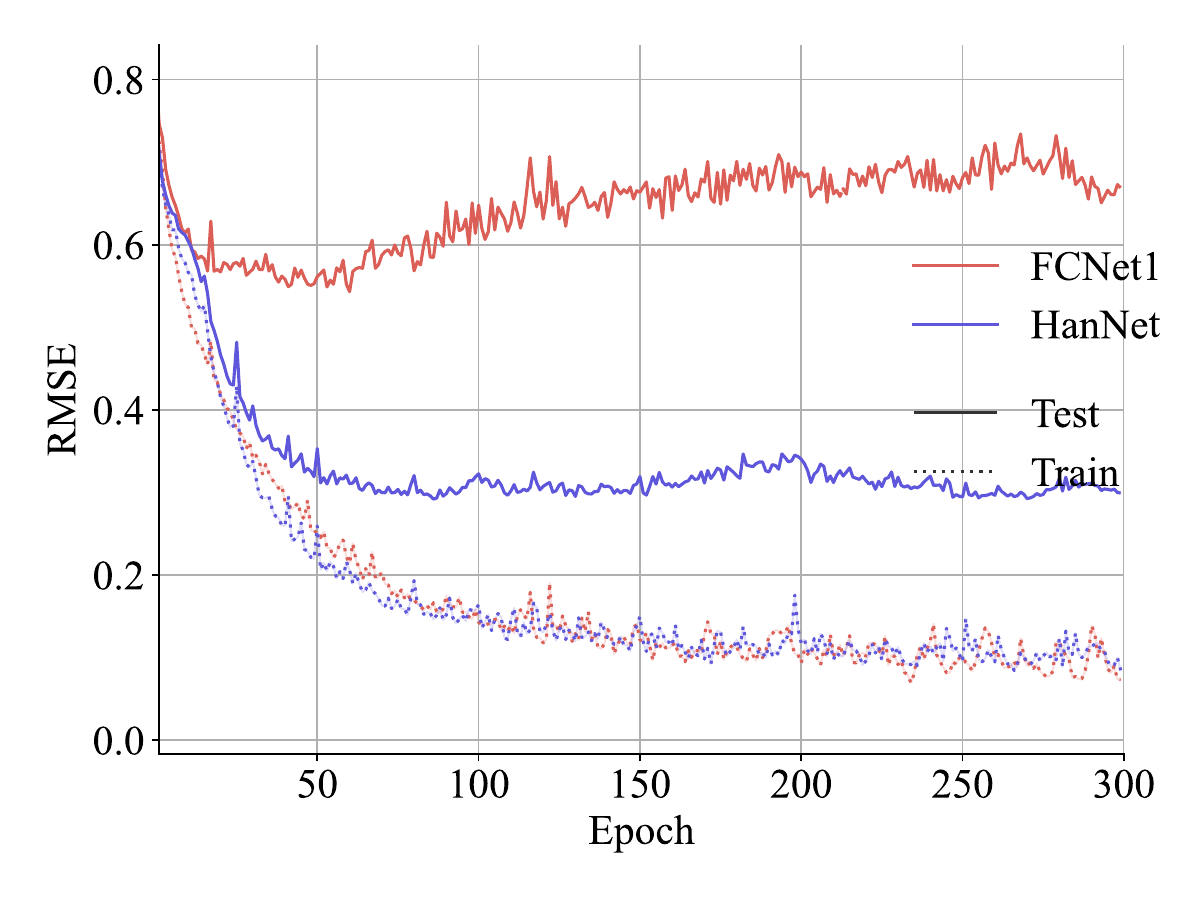}
	\includegraphics[width=.22\textwidth,trim=20 20 20 20, clip]{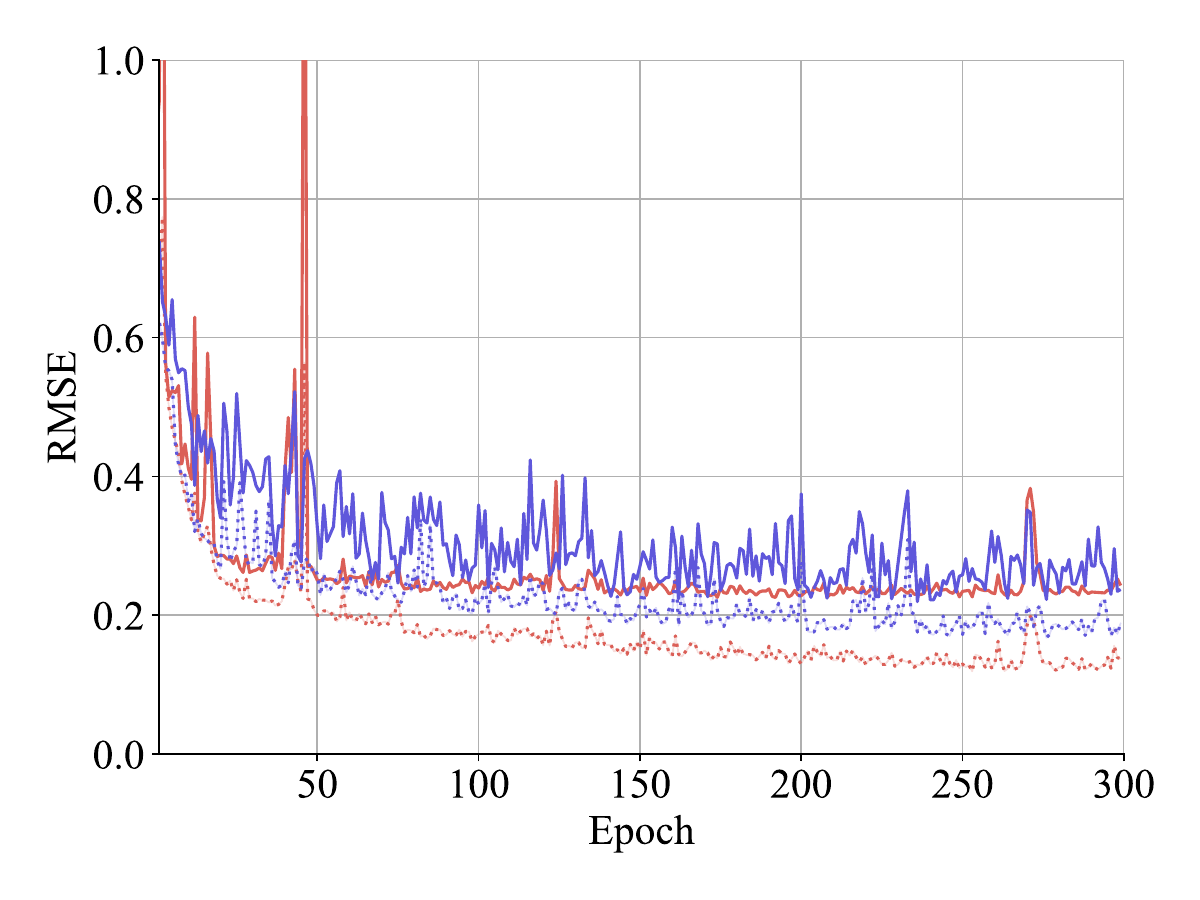}
	\caption{The training~(solid line) and testing~(dotted line) RMSE on Bikesharing. Left: $\delta=0.2$; right:  $\delta=0.8$. Red: for FCNet1; blue: for HanNet.}
	\label{fig:elevators}
\end{figure}

\subsection{Use of Han-layers in Image Classification}

{
In this section, we experiment on integrating Han-layers into existing models for image classification. First, we use Han-layers to replace some MLP-layers in MLP-Mixer models\cite{mixertolstikhin2021}, and observe the resulting model performances. Next, we extend our investigation to a lightweight model,  MobileViT~\cite{mehta2021mobilevit}.
}

\subsubsection{Using Han-layers
in MLP-Mixer}\label{H/M-mix}

\begin{figure}[ht]
	\vspace{-.2cm}
	\centering
	\includegraphics[width=.5\textwidth,trim=20 150 0 60, clip]{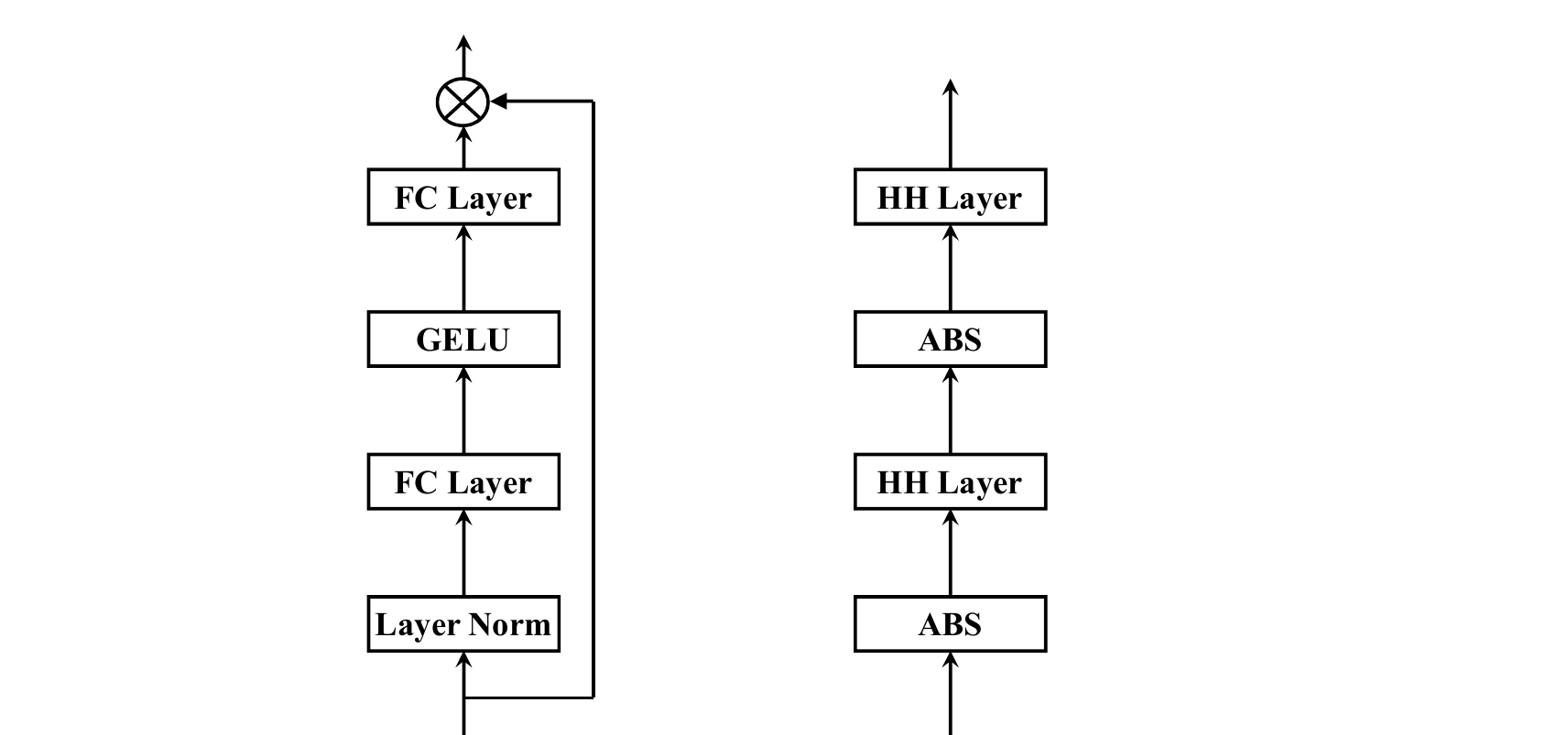}
	\caption{Two implementations of the channel mixing module using MLP-Mixer and Han-Mixer, respectively. HH denotes Householder.
	}
	\label{fig: two mixers}
	\vspace{-.2cm}
\end{figure}

In literature, the term Multi-Layer Perceptron (MLP) is often used exchangeably with FCNet. Recently, MLP-dominated models have seen a wave of revivals for image recognition tasks~\cite{mixertolstikhin2021, liu2021pay}. MLP-dominated models (without multi-head attention) are much more concise than Transformer-based models~\cite{dosovitskiy2020image, touvron2021training} but can still maintain test performances on very large-scale datasets. The motivation of MLP-Mixer~\cite{mixertolstikhin2021} is to use the purely fully connected layers to remove attention architectures. An MLP-Mixer block is the elementary unit in MLP-Mixer models that consists of several FC-layers and skip-connections to form the map from input $X$ to output $Y$ as is shown in Figure~\ref{fig: two mixers}.

\begin{figure*}[htb]
	
	\centering
	\includegraphics[width=1.\textwidth,trim=20 370 0 100, clip]{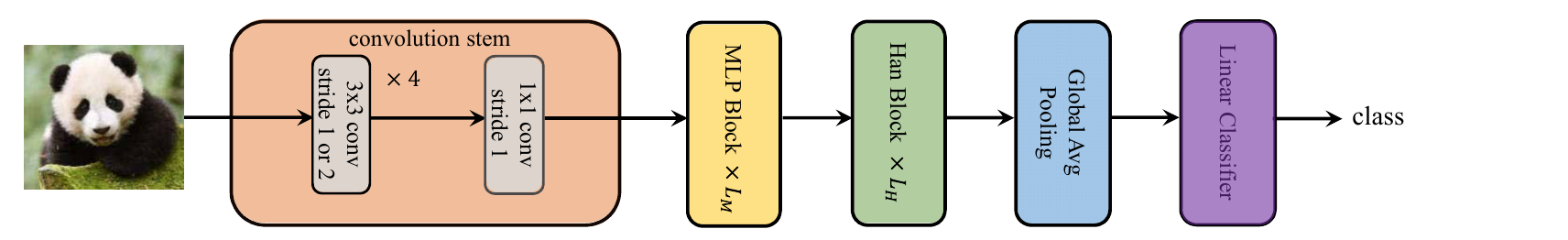}
	\caption{Overall structure of the tested Han/MLP-Mixer model ($L_M$ can be zero). }
	
	\label{fig:mixer}
	\vspace{-.2cm}
\end{figure*}

\begin{eqnarray}
	Z &=&X +\textbf{GELU}\left(W_2 \ \textbf{Layer Norm}(W_1 X) \right),  \label{eq:token-mix}\\
	Y &=&Z +\textbf{GELU}\left( \textbf{Layer Norm}(Z W_3) \ W_4 \right), \label{eq:channel-mix}
\end{eqnarray}
where the first-row \eqref{eq:token-mix} is called token-mixing for cross-token communication, and the second-row \eqref{eq:channel-mix} is called channel-mixing for cross-channel communication, both being of MLP structure. Here we form our Han-Mixer block by replacing all weight matrices $W_i$ by Householder matrices $H_i$, $i=1,2,3,4$, and all activation functions by the absolute function \textbf{ABS}, that is,
\begin{equation} \label{eq:han-mix}
	Z = \textbf{ABS}(H_2 \textbf{ABS}(H_1 X)),  ~~
	Y = \textbf{ABS}(\textbf{ABS}(Z H_3) H_4),
\end{equation}
where we remove skip connections and layer normalizations (since HanNets does not suffer from gradient problems). The resulting Han/MLP-Mixer models are shown in Figure~\ref{fig:mixer}, where we arrange some Han-Mixer blocks after MLP-Mixer blocks (which may be empty). The main reason for us to combine the two types of Mixer blocks is that, short of drastically increasing network width, Han-Mixers alone cannot always provide enough model parameters for large-scale datasets. In addition, we use the convolution stem recommended by~\cite{xiao2021early} instead of the one in~\cite{mixertolstikhin2021}.

\subsubsection{Settings}\label{sec: setting on image data}
\begin{table}[hptb]
	\caption{The architecture on one MLP-block or Han-block. For simplification, in our experiments, all hidden MLP weights are square matrices. }
	\smallskip
	\centering
	\resizebox{1.\columnwidth}{!}{
	\begin{tabular}{lccccc}
		\toprule[1.2pt]
		\multicolumn{1}{c}{} & \multicolumn{2}{c}{CIFAR}  & \multicolumn{2}{c}{STL10}  & \multicolumn{1}{l}{Downsampled-ImageNet} \\
		\midrule
		Patch size  & \multicolumn{2}{c}{$4\times 4$}     & \multicolumn{2}{c}{$8\times 8$}   & $4\times 4$ \\
		Sequence Length  & \multicolumn{2}{c}{64}    & \multicolumn{2}{c}{144}   & 64  \\
		\multicolumn{1}{l}{\multirow{1}{*}{Channel number}} & \multicolumn{2}{c}{512}  & \multicolumn{2}{c}{512} & 1024       \\
		\bottomrule[1.2pt]
	\end{tabular}
	}
	\label{tab:block-structure}
\end{table}
We summarize various configurations of each block on different data sets in Table~\ref{tab:block-structure}. We run Mixers using the following number of layers \{1, 2, 4, 8, 12, 16\} and select the best one on each dataset. We use the convolution stem recommended by~\cite{xiao2021early} instead of the one in~\cite{mixertolstikhin2021}. Our convolutional stem designs use five layers, including four $3\times 3$ convolutions and a single $1\times 1$ final layer. The output channels are [64, 128, 256, 512, 512] on CIFAR10, CIFAR100 and STL10, and [128, 256, 512, 1024, 1024] on Downsampled-ImageNet. Some data augmentation methods are used, such as random HorizontalFlip, random augmentation\cite{ cubuk2020randaugment}, and Cutout\cite{devries2017improved}.
For these datasets, we choose a modified Adam called AdamW~\cite{loshchilov2018decoupled}. Detailed settings are in Table~\ref{tab:adamimage} below. 

\begin{table}[ht]
	\caption{Dataset statistics and optimizer settings. $N$ is the number of samples, and Dim is the dimension of data vectors.}
	\smallskip
	\centering
	\resizebox{1.\columnwidth}{!}{
		\begin{tabular}{lcccccc}
			\toprule[1.2pt]
			Dataset & Dim & $N$ & LR  &  weight decay &batch size  &  epochs \\ \midrule
			CIFAR10/ CIFAR100 & $3\times32\times32$ & 60000  & \multirow{3}{*}{0.001} & \multirow{3}{*}{0.1} & 256 & 600\\
			STL10  & $3\times 96\times 96$  & 13000   &  & &   64  & 300\\
			Downsampled-ImageNet & $3\times32\times32$  & 1331167 &  & &  512  & 300 \\
			\bottomrule[1.2pt]
		\end{tabular}
	}
	\label{tab:adamimage}
\end{table}

\subsubsection{Results}

In this section, we investigate HanNets' generalization ability in image classification by comparing MLP-Mixers with our Han/MLP-Mixers presented in Subsection~\ref{H/M-mix}. The comparisons are made with 3 standard image datasets and a down-sampled version of ImageNet (due to computing source limitations).

Empirical evidence from MLP-based models~\cite{mixertolstikhin2021} suggests that these models can achieve the state-of-the-art results on very large-scale datasets. Since the datasets used in our tests are not large enough to play to the full strengths of Mixer models, our aim here is not to observe how close our results can approach the state-of-the-art, but how Han-layer structures can impact the performance of Mixer models. 

\begin{table}
	\caption{Error rates~(\%) on four datasets: STL10, CIFAR10, CIFAR100 and Downsampled-ImageNet.} 
	\smallskip
	\centering
	\resizebox{.9\columnwidth}{!}{
		\begin{tabular}{lccccc}
			\toprule[1.2pt]
			& \#Param    & STL10 & CIFAR10 &  CIFAR100  \\
			\midrule
			CNN stem & 1.82 M   & 18.2     & 7.2    &  27.8          \\
			+MLP~(2)  &  2.89 M & 18.2      &  5.3      &  27.2    \\
			+MLP~(4) & 3.96 M  & 18.7    & 5.5       & 27.1      \\
			+MLP~(0) + Han~(16) & 1.86 M & \textbf{15.3}    & 5.9       & 26.7       \\
			+MLP~(1) + Han~(16) &  2.39 M &  16.7    &  5.0    &  24.6    \\
			+MLP~(2) + Han~(16) &  2.93 M &  17.1    &  4.7    &  \textbf{24.4}    \\
			+MLP~(4) + Han~(16) &  4.00 M & 17.8    & \textbf{4.6}    & 24.7   \\
			\midrule
			ImageNet-Downsampled & \multicolumn{2}{c}{\#Param}  & Top1 error &  Top5 error \\
			\midrule
			CNN stem  & \multicolumn{2}{c}{8.28 M}  &  55.7     & 31.2              \\
			+MLP~(4) &  \multicolumn{2}{c}{16.7 M} & 44.2    & 21.2   \\
			+MLP~(8) & \multicolumn{2}{c}{27.2 M} & 42.6    & 20.2   \\
			+MLP~(0) + Han~(16) & \multicolumn{2}{c}{8.35 M}  &  48.9    &  26.1   \\
			+MLP~(4) + Han~(16) & \multicolumn{2}{c}{16.8 M} & \textbf{41.1}    & \textbf{19.2}   \\
			WideResNet~\cite{chrabaszcz2017downsampled} & \multicolumn{2}{c}{37.1 M}  & \textbf{41.0}    & \textbf{18.9}    \\
			\bottomrule[1.2pt]
		\end{tabular}
	}
	\label{tab:Downsampled-ImageNet}
	\vspace{-.2cm}
\end{table}

Table~\ref{tab:Downsampled-ImageNet} reports test results from various MLP-Mixer and Han/MLP-Mixer models. On the smaller dataset STL10, HanMixers alone can already produce the best result. On larger datasets CIFAR10 and CIFAR100, enhanced performance from HanMixers is achieved either by adding a relatively few parameters (e.g., MLP~(2) or (4) vs. MLP~(2) or (4) + Han~(16)), or even with far fewer parameters (e.g., MLP~(4) vs. MLP~(1)+Han~(16)). On Downsampled-ImageNet, our Han/MLP-Mixer combination model clearly outperforms pure MLP-Mixers and offers a competitive performance with WideResNet~\cite{chrabaszcz2017downsampled}~(41\% Top1 error) while using only 40\% parameters.  In summary, the benefits of using Han-Mixers should be empirically evident according to these experiments. 

\subsection{Using Han-layers in MobileViT}

{In this section, we incorporate Han-layers into a lightweight vision transformer model, MobileViT~\cite{mehta2021mobilevit}, and compare model performance after MLP-layers are replaced by Han-layers.  MobileViT combines convolutional and fully connected layers to achieve good performance while being lightweight. Details of this model, which are beyond the scope of this paper,  can be found in \cite{mehta2021mobilevit}.

In MobileViT~\cite{mehta2021mobilevit}, an MLP-block is of the form
\[
	Y = X +\textbf{Swish} \left( \textbf{Layer Norm}(X) W_1 \right) W_2,
\]
which uses a layer normalization and Swish activation~\cite{elfwing2018sigmoid}.
The above MLP-block is replaced, one-on-one, by the Han-block
\[
 Y = \textbf{ABS}(\textbf{ABS}(X H_1) H_2),
\]
where $H_1$ and $H_2$ are Householder matrices.

We experiment on two variants of the MobileViT model: MobileViT-XXS and MobileViT-XS with different sizes (see \cite{mehta2021mobilevit}).  The training settings are similar to those in the previous MLP-Mixer experiments, except that we resize each image to $3\times 256\times 256$ according to the settings of MobileViT. Detailed settings are given in Table~\ref{tab:settings-mobilevit}.

\begin{table}[ht]
	\caption{Experiment settings on MobileViT.  Dim represents data dimension.}
	\smallskip
	\centering
	\resizebox{1.\columnwidth}{!}{
		\begin{tabular}{cccccc}
			\toprule[1.2pt]
			Dataset & Dim & LR  &  weight decay &batch size  &  epochs \\ \midrule
			CIFAR10/100 & $ \multirow{2}{*}{$3\times 256\times 256$}$ & \multirow{2}{*}{0.001} & \multirow{2}{*}{0.1} & 128 & \multirow{2}{*}{300}\\
			STL10  &      &  & &   64  & \\
			\bottomrule[1.2pt]
		\end{tabular}
	}
	\label{tab:settings-mobilevit}
\end{table}

Testing performances of using MLP-blocks and Han-blocks are presented in Table~\ref{tab:Mobile-ViT}. Across the three datasets (STL10, CIFAR10 and CIFAR100) and two model variants, the use of Han-blocks reduces the model parameter counts by approximately one half, while yielding comparable or better performance. These empirical findings again strongly support the use of Han-layers under suitable conditions.

\begin{table}[htp]
\caption{Error rates~(\%) on datasets STL10, CIFAR10 and CIFAR100, for models MobileViT-XXS and MobileViT-XS.}
	\smallskip	
	\centering
\resizebox{.9\columnwidth}{!}{
\begin{tabular}{llcccc}
    \toprule[1.2pt]
\multicolumn{1}{l}{MobileViT} & Block & \# Param & STL10 & CIFAR10 & CIFAR100 \\
\midrule
\multirow{2}{*}{XXS} & MLP & 1.01 M & 12.4 & \textbf{4.4} & 22.7 \\
 & Han & 0.55 M & \textbf{12.0} & 4.5 & \textbf{22.5} \\
 \midrule
\multirow{2}{*}{XS} & MLP & 2.00 M & 11.5 & 4.2 & 21.1 \\
 & Han & 0.97 M & \textbf{11.0} & \textbf{3.9} & \textbf{20.4} \\
\bottomrule[1.2pt]
\end{tabular}
}
\label{tab:Mobile-ViT}
\end{table}

\section{Conclusions}

We propose a lightweight layer structure (called Han-layer) for neural networks that has guaranteed gradient stability and 1-Lipschitz continuity.  We provide extensive empirical evidence demonstrating the efficacy of Han-layers in reducing model parameters in certain neural networks. That is, when used strategically as substitutes for fully connected layers, a relatively small number of Han-layers can maintain or improve generalization performance.  Moreover, we observe impressive generalization capabilities of HanNets on some synthetic datasets.   These proof-of-concept results suggest that Han-layer can potentially become a useful building block in constructing powerful deep neural networks.  Of course, further investigations, especially theoretical ones, are still needed to fully assess the strength and weakness of Han-layers.


Since Householder matrices are square, as it is currently defined, a Han-layer cannot change dimensionality. Moreover, being light-weighted, Han-layers alone may not effectively provide enough parameters that are needed for building large models. Due to such limitations, Han-layers should be used strategically in combination with fully connected layers, and other suitable layer structures, to construct lean and effective deep neural networks in applications.


\section{Acknowledgments}

The work is supported by Shenzhen Research Institute of Big Data, Shenzhen Science and Technology Program (Grant Number GXWD20201231105722002-20200901175001001) and NSFC 12141108.


\bibliographystyle{elsarticle-num}
\bibliography{ref}

 \appendix
\section{Other Figures on Regression Datasets}

\begin{figure}[th]
	\centering
	\subfigure[Bikesharing]{
		\includegraphics[width=.22\textwidth,trim=0 0 0 0, clip]{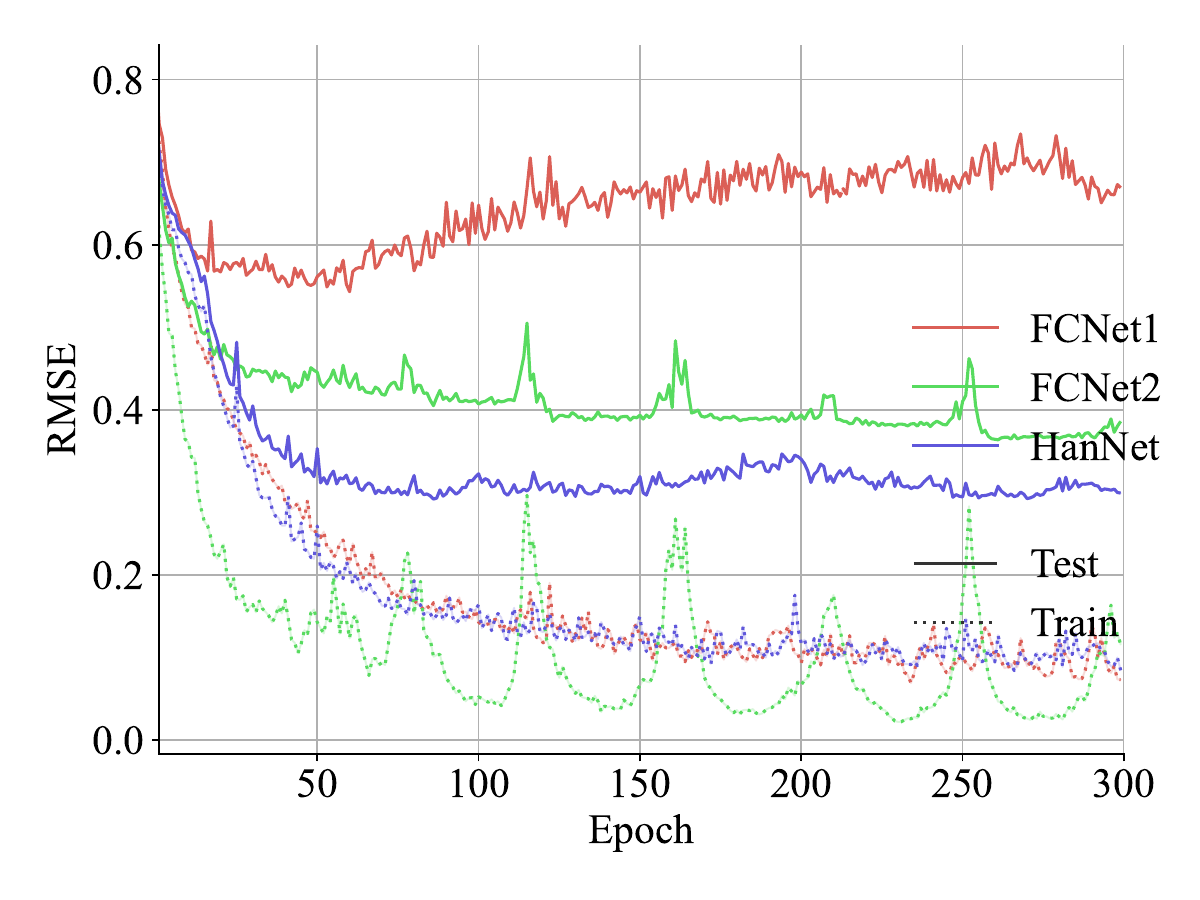}
		\includegraphics[width=.22\textwidth,trim=0 0 0 0, clip]{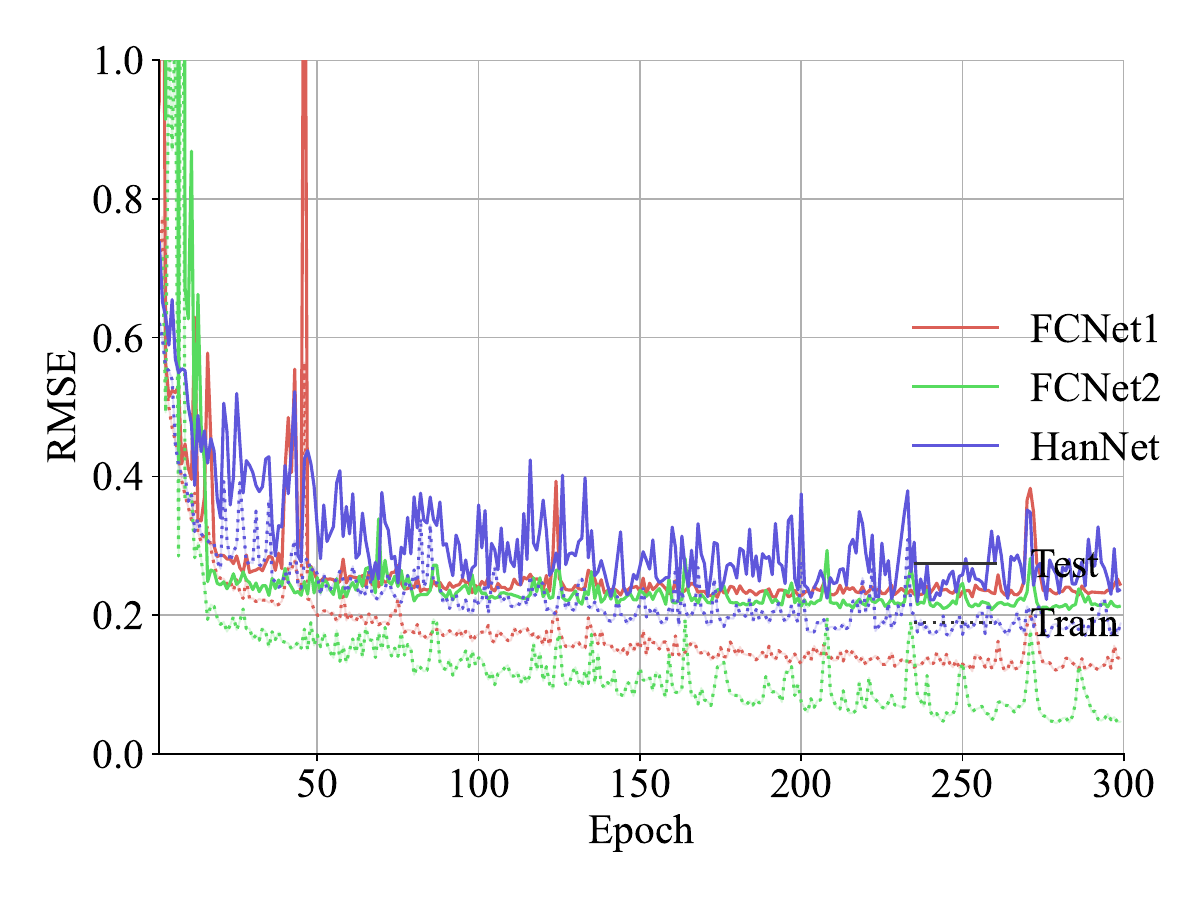}}
	\subfigure[Calhousing]{
		\includegraphics[width=.22\textwidth,trim=20 20 20 20, clip]{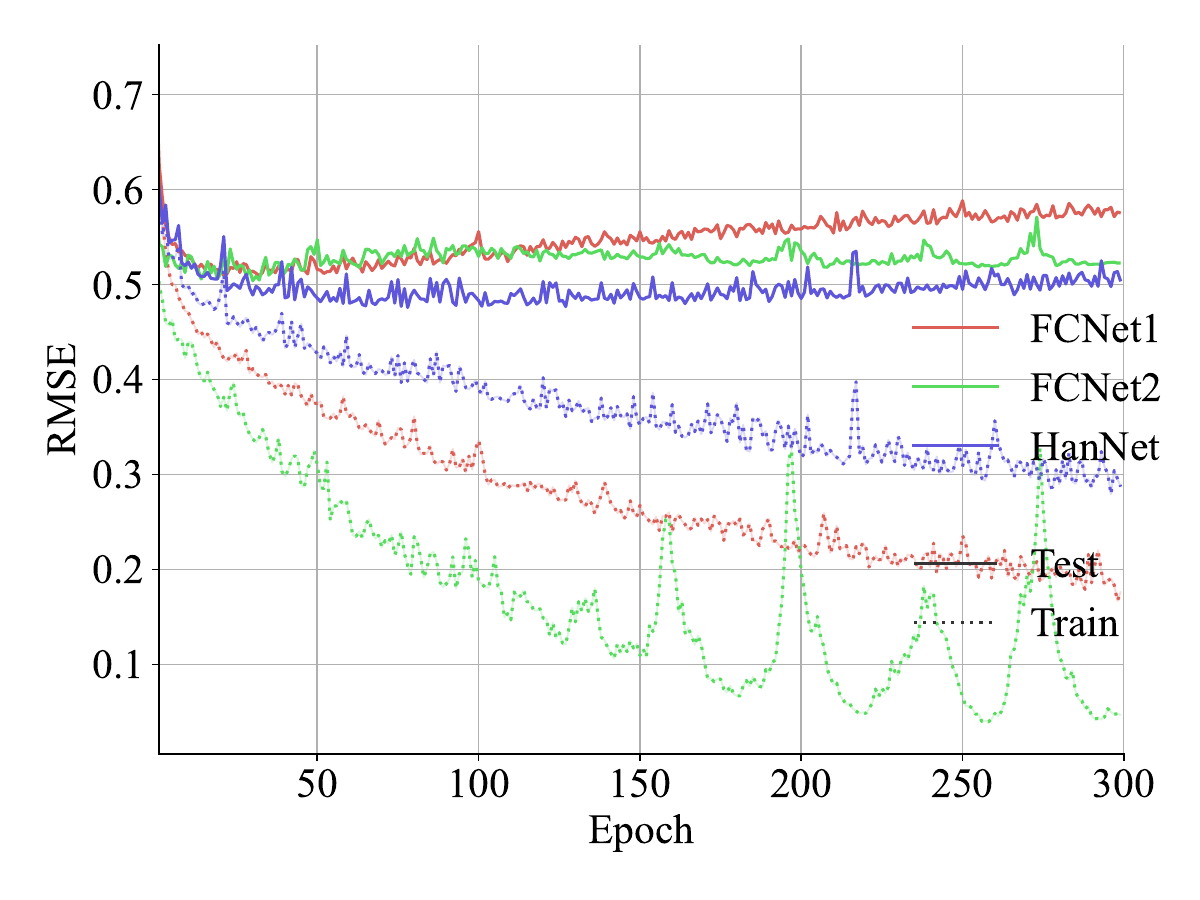}
		\includegraphics[width=.22\textwidth,trim=20 20 20 20, clip]{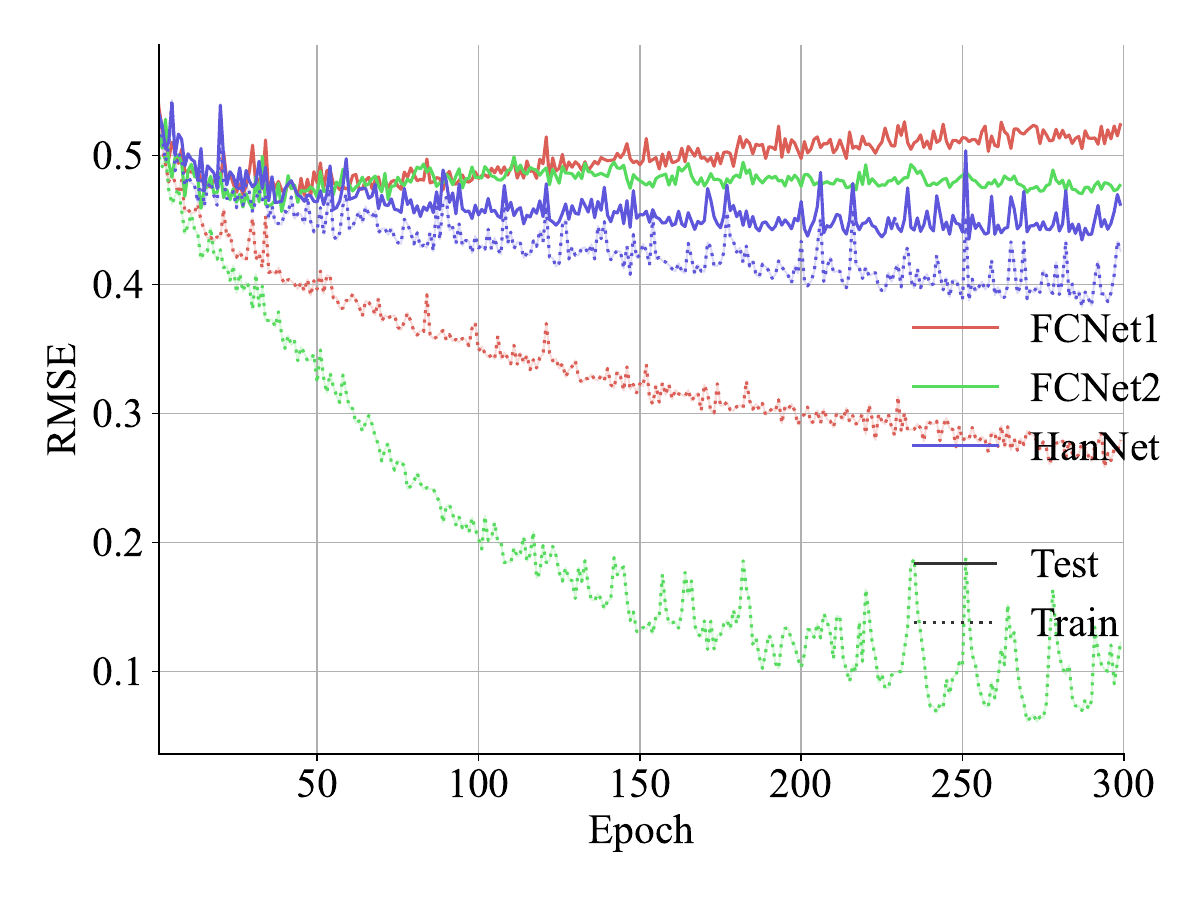}}	
	\subfigure[Elevators]{
		\includegraphics[width=.22\textwidth,trim=20 20 20 20, clip]{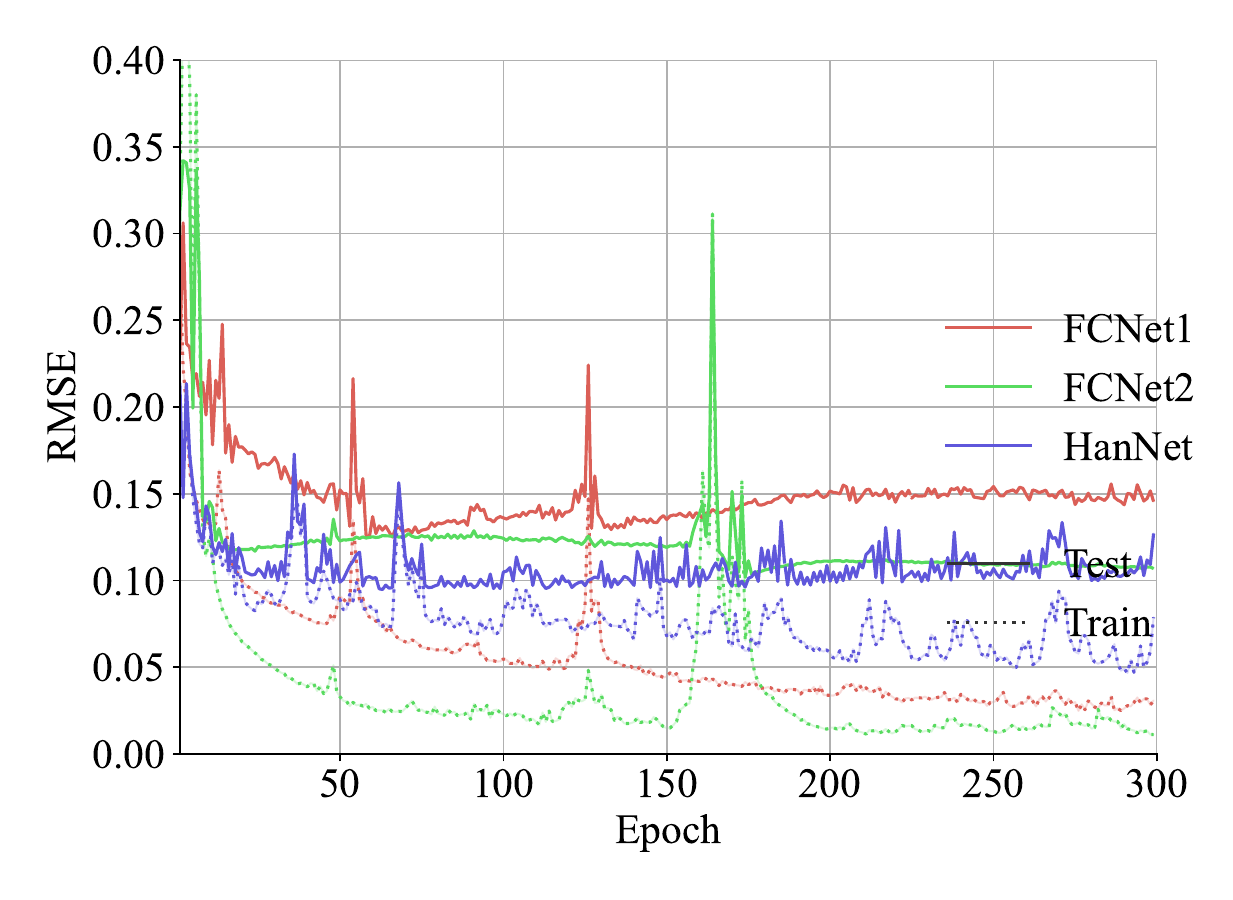}
		\includegraphics[width=.22\textwidth,trim=20 20 20 20, clip]{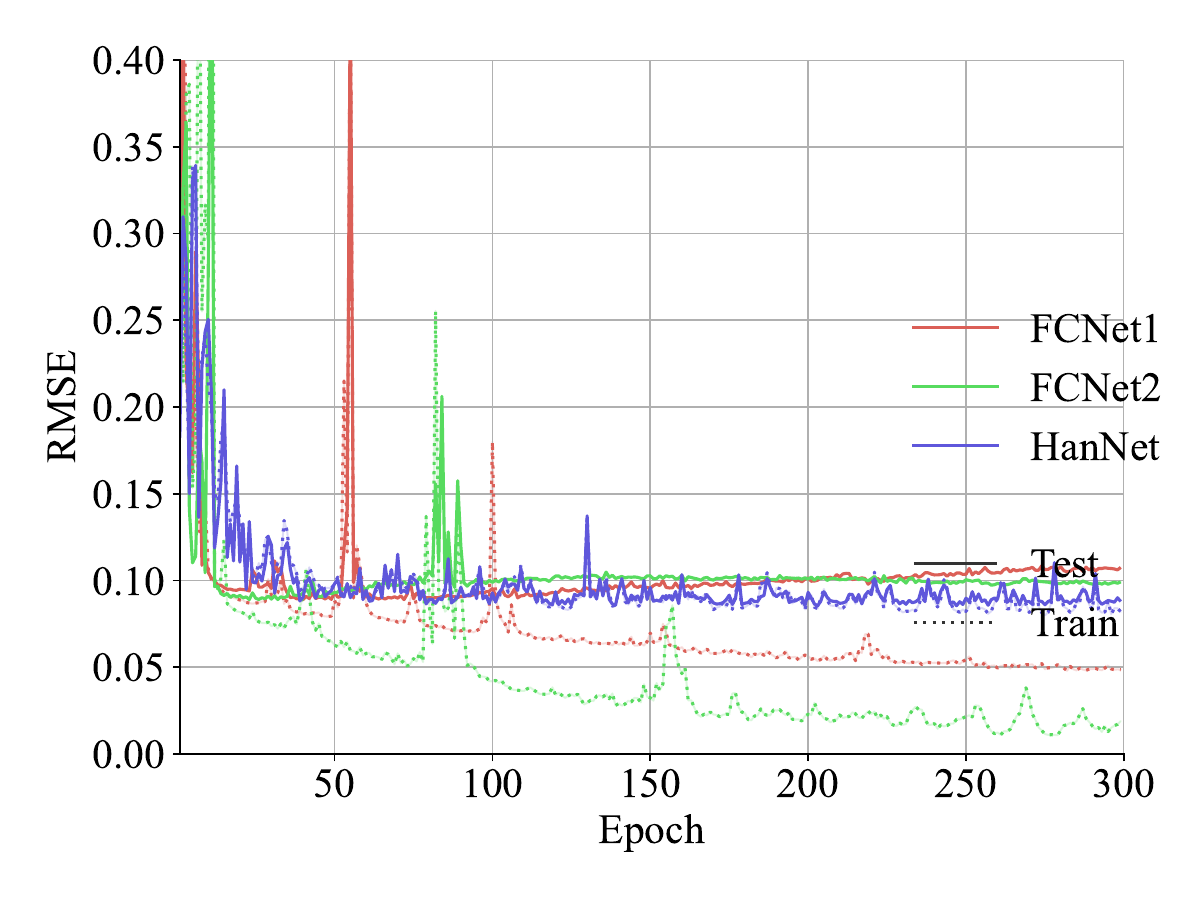}}	
	\subfigure[Skillcraft]{
		\includegraphics[width=.22\textwidth,trim=20 20 20 20, clip]{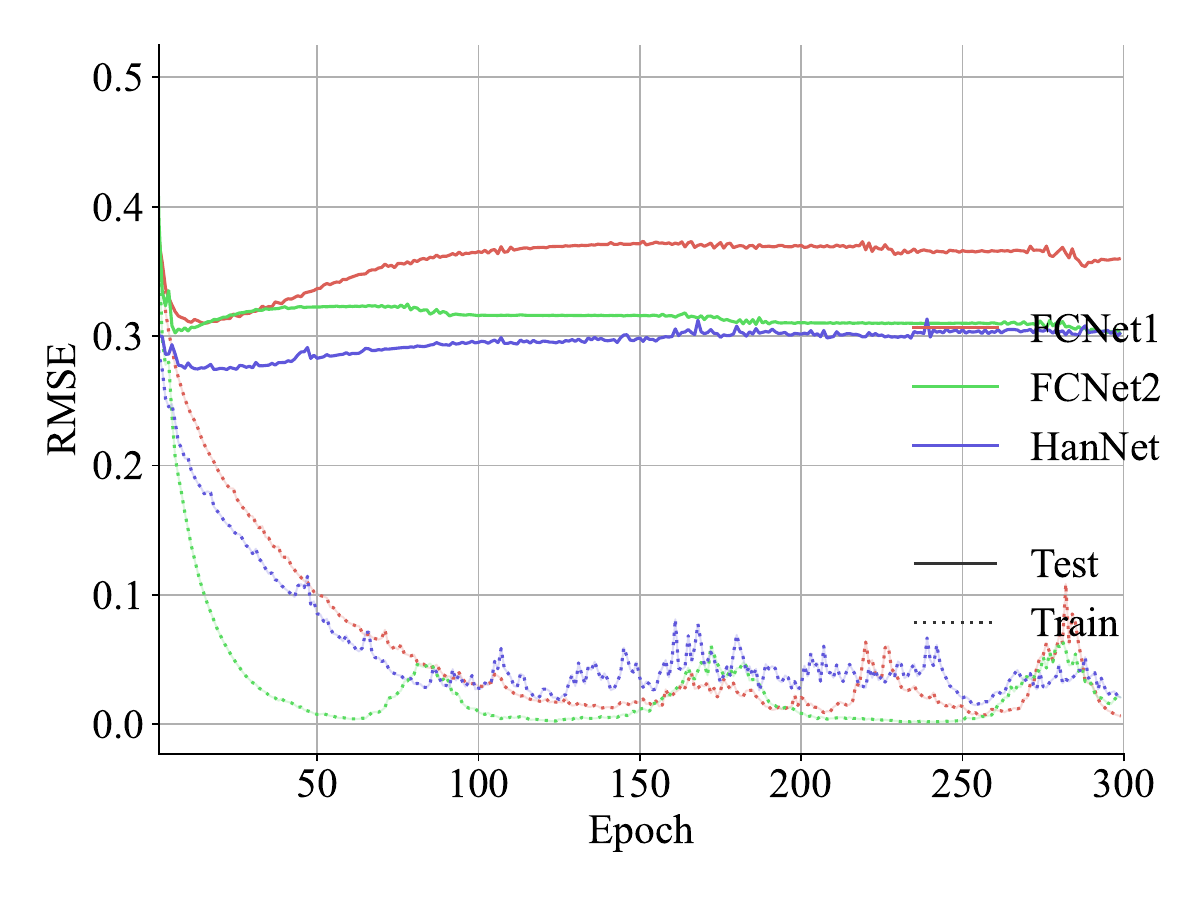}
		\includegraphics[width=.22\textwidth,trim=20 20 20 20, clip]{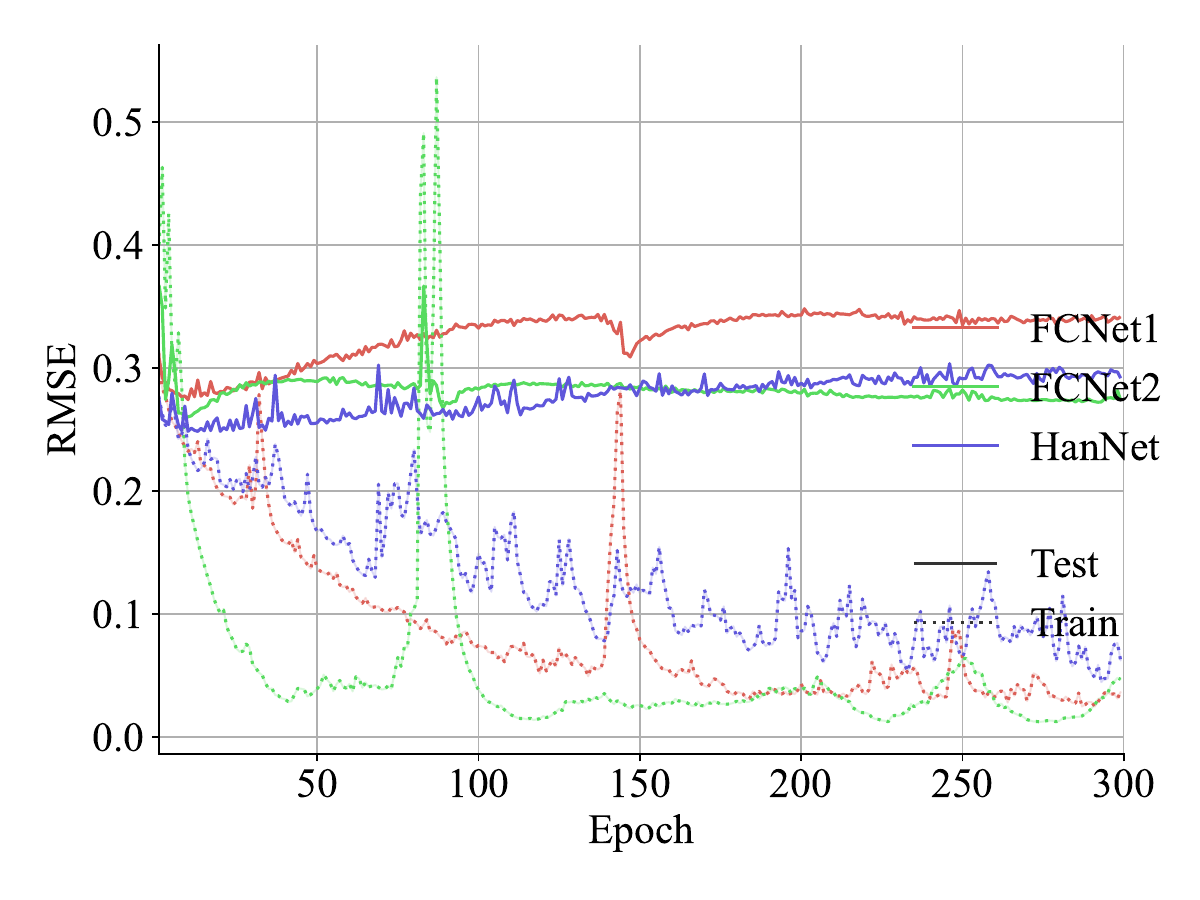}}	
	\subfigure[Parkinsons]{
		\includegraphics[width=.22\textwidth,trim=20 20 20 20, clip]{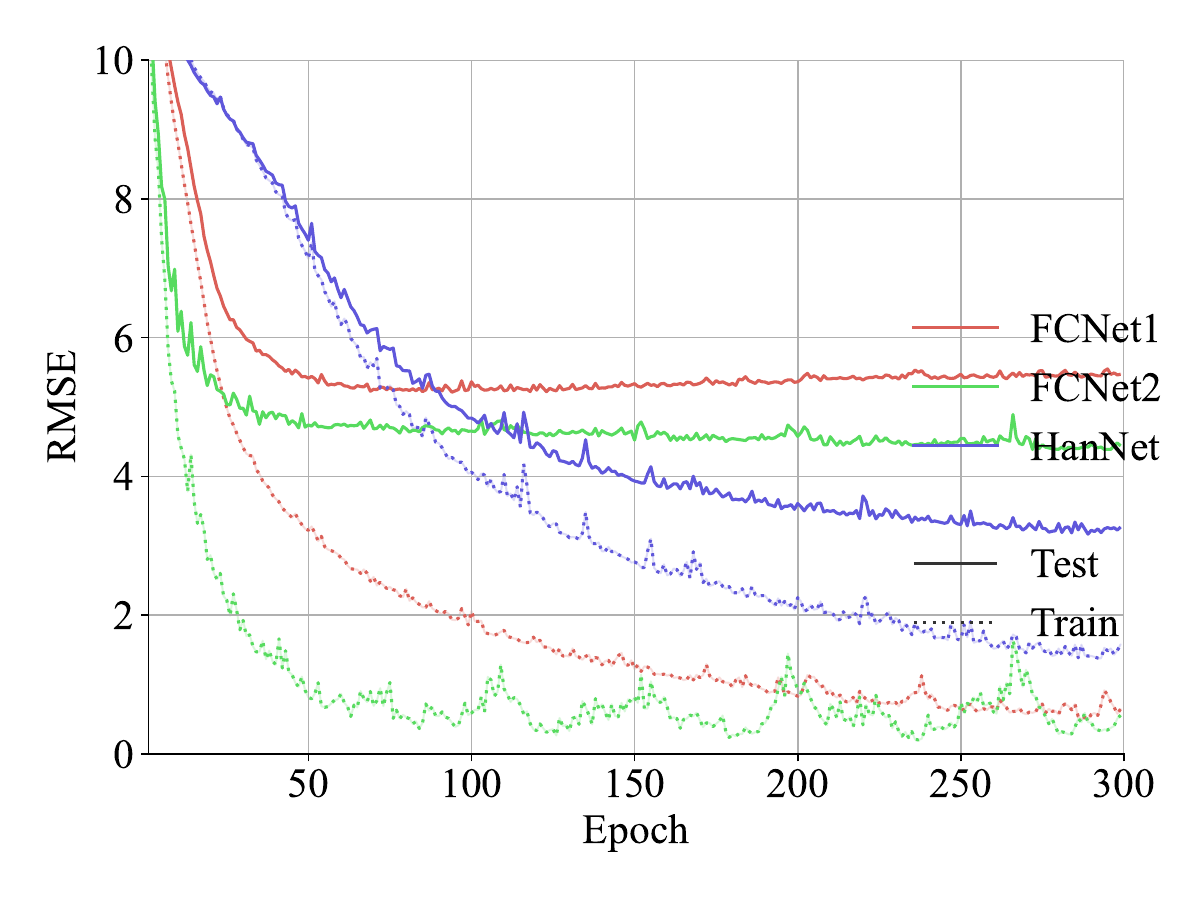}
		\includegraphics[width=.22\textwidth,trim=20 20 20 20, clip]{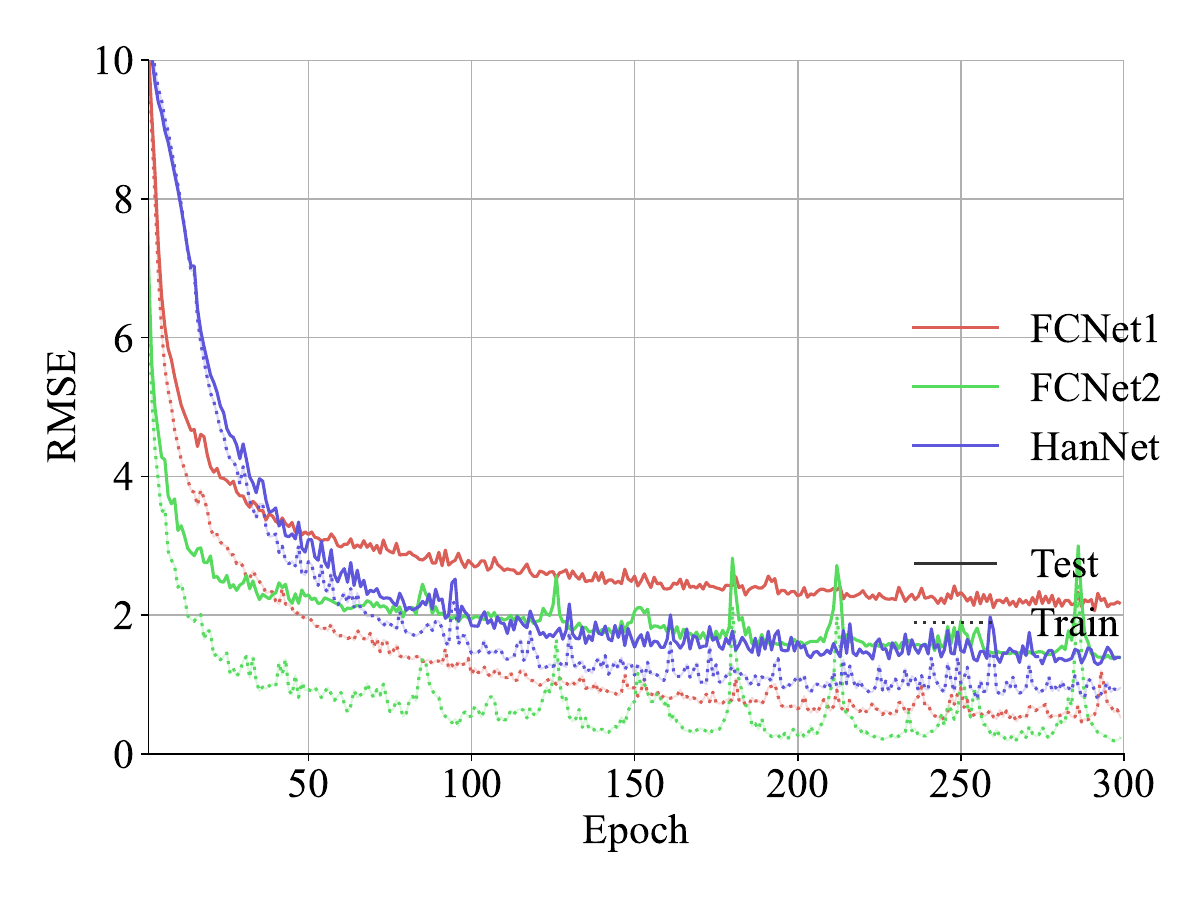}}
	
	\caption{The train~(solid line) and test~(dotted line) RMSE on 5 regression datasets. Left: $\delta=0.2$; right:  $\delta=0.8$. Red: for FCNet1; green: for FCNet2; blue: for HanNet.}
\end{figure}




\end{document}